\documentclass[journal]{IEEEtran}
\usepackage{amsmath,amsfonts}
\usepackage{algorithmic}
\usepackage[colorlinks=true, citecolor=blue, linkcolor=red, urlcolor=blue]{hyperref}
\usepackage{array}
\usepackage[caption=false,font=normalsize,labelfont=sf,textfont=sf]{subfig}
\usepackage{textcomp}
\usepackage{stfloats}
\usepackage{url}
\usepackage{verbatim}
\usepackage[capitalize]{cleveref}
\usepackage{graphicx}
\usepackage{amsthm}
\usepackage{amssymb}
\usepackage[left=1.25cm,right=1.2cm,top=1.3cm,bottom=1.2cm]{geometry} % 调整页边距
\newtheorem{theorem}{Theorem}[section]
\newtheorem{lemma}{Lemma}[section]
\newtheorem{corollary}{Corollary}[section]
\newtheorem{definition}{Definition}[section]
\def\BibTeX{{\rm B\kern-.05em{\sc i\kern-.025em b}\kern-.08em
    T\kern-.1667em\lower.7ex\hbox{E}\kern-.125emX}}
\usepackage{balance}
\begin{document}
\title{PSF-Informed Reconstruction: An Enhanced 2D Directional Filtered Backprojection Method for Cone-Beam Rotary Computed Laminography}
\author{Chengxiang Wang, Wei Yu, and Baodong Liu
\thanks{Chengxiang Wang is with the School of Mathematical Sciences, Chongqing Normal University, Chongqing, 401331, China (Email:20190009@cqnu.edu.cn).\\ \indent Wei Yu is with the School of Biomedical Engineering and Imaging, Xianning Medical College, Hubei University of Science and Technology, Xianning, 437100, China (Corresponding author, Email:cq\_yuwei@163.com).\\ 
\indent Baodong Liu is with the Beijing Engineering Research Center of Radiographic Techniques and Equipment, Institute of High
Energy Physics, Chinese Academy of Sciences, Beijing, China; the School of Nuclear Science and Technology, University of Chinese Academy of Sciences, Beijing, China; and the Jinan Laboratory of Applied Nuclear Science, Jinan, China (Email: liubd@ihep.ac.cn). \\
\indent This work was supported in part by the National Natural Science Foundation of China (No.62371184);  the Natural Science Foundation of Chongqing (Nos.CSTB2023NSCQ-MSX0148 and CSTB2023NSCQ-LZX0054);  the Scientific and Technological Research Program of the Chongqing Municipal Education Commission (Nos.KJQN202300535 and KJZD-K202300506) and Xianning Science and Technology Talent Service Special Project(2024RCFW007). }}

\markboth{IEEE TRANSACTIONS ON IMAGE PROCESSING,}%
{PSF-Informed Reconstruction: An Enhanced Directional Filtered Backprojection Method for Cone-Beam Rotary Computed Laminography}

\maketitle

\begin{abstract}
Cone-beam rotary computed laminography (CL) presents a viable alternative to cone-beam CT for imaging objects with high aspect ratios. Conventional reconstruction techniques, such as CL-FBP and CL-FDK, often yield suboptimal results characterized by pronounced superimposition artifacts, reduced contrast, and blurred edges, primarily due to their reliance on 1D row-wise filtering along detector rows. To address these limitations and improve reconstruction quality, we proposed  a novel  2D directional filtered backprojection (DFBP) method based on the point spread function (PSF). First, an exact expression for the projection process is derived by establishing the relationship between  a reconstructed pixel and its corresponding projection point on the detector. Second, a revised  backprojection process incorporating two weighting factors is introduced to counteract the nonlinearity inherent in conventional backprojection, leading to a simplified PSF formulation. Third, to circumvent the computational burden of 3D image space filtering, an approximate method by linearizing the coordinate relationship in Fourier space under some assumptions is developed that transfers the 3D filtering operation to the 2D projection space. This approach facilitates the derivation of a formally simple yet effective 2D directional filter, which integrates a 2D von Hann window and a 2D sinc window.  Experimental results demonstrate that the proposed method produces reconstructions with superior contrast, sharper boundaries, and fewer superimposition artifacts compared to CL-FBP and CL-FDK. Moreover, it enables more accurate recovery of internal structures, such as bubble defects and electronic components.
\end{abstract}


\section{Introduction}
\IEEEPARstart{C}{one-beam} computed tomography (CT) is extensively employed in non-destructive evaluation due to its capacity to generate high-resolution internal images\cite{fisher2019laminography}. Nevertheless, its conventional scanning geometry is ill-suited for high-aspect-ratio objects, which  are difficult to penetrate by X-rays along their longitudinal axis, resulting in artifacts in the reconstructed images\cite{myagotin2013efficient}. Typical examples include printed circuit boards (PCBs)\cite{shi2023automatic,gui20233}, electric vehicle battery packs\cite{lin2022design}, and insulated gate bipolar transistors (IGBTs)\cite{wei2017micro}. As a viable alternative, cone-beam rotary computed laminography (CL) addresses this limitation by tilting either the X-ray source and detector assembly in a conventional cone-beam CT system or the object being scanned\cite{Wang2025,van2016fast,zhao2021fast,lu2023anisotropic,samber2021flexct}. System tilting maintains a consistent X-ray transmission profile across all projection angles and extends the detector's effective field of view to accommodate larger objects. However, the titled geometry of cone-beam rotary CL  leads to incomplete projection data and a loss of interlayer information from the scanned object, thereby further ill-posing the reconstruction problem\cite{lu2023anisotropic,liu2023cone}. This results in the mixing of information between adjacent slices, giving rise to laminography-specific artifacts in the reconstructed image, such as superimposition artifacts, which ultimately  degrade the image quality\cite{fisher2019laminography,lu2023anisotropic,liu2023cone,sun2021reconstruction}. \\
\indent To mitigate superposition artifacts, the incorporation of additional information—such as knowledge of a finite support or constraints on the function's value range—is required. Iterative reconstruction methods offer considerable flexibility in this respect, as they can be readily adapted to non-standard acquisition geometries and allow for the integration of various forms of prior knowledge.   The simultaneous iterative reconstruction technique (SIRT)\cite{gilbert1972iterative} is capable of producing high-quality reconstructed images when applied to complete projection data with a high signal-to-noise ratio. However, when reconstructing objects with a large aspect ratio from truncated projections, SIRT tends to generate elevated values near the image boundaries while suppressing values in interior regions. This imbalance leads to diminished contrast and introduces artifacts within the region of interest (ROI). To mitigate ROI artifacts, a ray-length-based correction method is proposed by assigning a distinct weight to each ray\cite{schorr2014ray,WANG2025103396}. Prior knowledge of the object's shape—such as CAD or STL data—or image sparsity is incorporated into the reconstruction model to  reduce ROI artifacts\cite{schorr2014ray,Gui2023,GONDROM1999141}. Several regularization techniques have been incorporated into the reconstruction model, including total variation (TV)\cite{gui20233}, 
weighted total variation (WTV)\cite{tang2024cone}, and $l_{0}$  minimization\cite{yu2017edge,LU2023102898}, among others. While iterative reconstruction methods can provide superior image quality, they are still far too slow and memory-intensive for routine use.\\
\indent Analytical reconstruction algorithms have gained popularity as a practical reconstruction method in real-world applications, owing to their relatively low computational complexity and acceptable image quality.  Circular tomosynthesis is a form of parallel-beam CL. In an effort to address this issue, Matsuo et al.\cite{matsuo1993three} proposed a reconstruction method based on inverse filtering, in which the point-spread function (PSF) of the projection and backprojection process was first derived, followed by the application of inverse filtering for image reconstruction. However, a significant limitation of this approach is its dependence on three-dimensional (3D) filtering in the image space, which is computationally intensive and time-consuming. In subsequent work, Lauritsch et al.\cite{lauritsch1998theoretical} developed a theoretical framework for filtered backprojection(FBP) in tomosynthesis, incorporating two additional window functions—von Hann and sinc—into the 3D filtering scheme established by Matsuo et al.  Helfen et al.\cite{helfen2011implementation}  proposed a FBP method for parallel-beam CL, employing the conventional one-dimensional(1D) Ram-Lak filter—a ramp function multiplied by a low-pass step function. Similarly,  Myagotin et al.\cite{myagotin2013efficient} developed a FBP algorithm for parallel-beam CL based on the Fourier slice theorem, which was implemented on multi-core clusters to accelerate the processing of discrete data, with the conventional 1D Ram-Lak filter employed in the method.  For cone-beam  rotary CL, Yang et al.\cite{yang2010new} and Que et al.\cite{que2012computed} proposed a CL-FBP method in which projection data are filtered row-by-row using a conventional 1D ramp filter before backprojection. However, this approach does not account for ray weights, leading to superimposition artifacts in the reconstructed images\cite{lin2022design,o2016recent}. Moreover, the filtering strategy lacks theoretical justification, and the origin of these artifacts remains unexplained. Voropaev et al.\cite{voropaev2016direct}  introduced a direct Fourier inversion reconstruction method for the CL scanning geometry. The continuous reconstruction formula was first derived and then discretized using 3D sinc interpolation. Furthermore, a projection resampling strategy was proposed, which reduces the computationally expensive 3D interpolation to a more efficient 1D operation, in which filtering is implemented with the standard 1D Ram-Lak filter. Lin et al.\cite{lin2022design} proposed a CL-backprojection filtering(CL-BPF) method for cone-beam rotary CL, in which the image is initially reconstructed via backprojection, followed by the application of a two-dimensional (2D) filter to each slice of the reconstructed volume. The Feldkamp-Davis-Kress (FDK) algorithm \cite{Buzug2008} is an approximate FBP method widely employed in circular cone-beam CT. It requires, however, that the central X-ray beam remains perpendicular to the detector plane and is free from tilt\cite{fisher2019laminography,lu2023anisotropic}. The direct application of the FDK algorithm to  cone-beam rotary CL (CL-FDK) may introduce certain artifacts in the reconstructed images. An alternative approach for reconstructing cone-beam rotary CL data involves mapping the projection data from the physical detector onto a virtual tilted detector, followed by reconstruction using the standard FDK algorithm. A key limitation of this method, however, is that it inherently introduces interpolation errors\cite{fisher2019laminography}.  Sun et al.\cite{sun2021reconstruction} proposed a cone-beam rotary CL reconstruction algorithm that employs a projection transformation. In this approach, the acquired projection data are first transformed into a virtual  data compatible with the FDK method. The image is then seamlessly reconstructed using the standard FDK method, utilizing the conventional 1D ramp filter. \\
\indent Due to its fast reconstruction speed and acceptable image quality, analytical reconstruction algorithms are well-suited for practical commercial CL applications. This paper will therefore focus on analytical reconstruction algorithms. Existing analytical methods for cone-beam rotary CL reconstruction typically employ the  1D ramp filter derived from traditional CT or parallel-beam CL during the projection filtering step. In parallel-beam CL, the relationship between a reconstructed pixel and its corresponding projection point on the detector is linear; hence, the 1D ramp filter—designed based on the point spread function(PSF)—can effectively suppress artifacts.  However, in cone-beam rotary CL, the mapping between a reconstructed pixel and its projection on the detector becomes nonlinear. Consequently, directly applying the conventional 1D ramp filter without designing a dedicated filtering strategy fails to adequately suppress artifacts and may even introduce distortions in the reconstructed object. To address these limitations and improve reconstruction quality, we proposed an approximate 2D directional filtered backprojection (DFBP) method based on the PSF.  First, an exact expression for the projection process is derived by establishing the relationship between  a reconstructed pixel and its corresponding projection point on the detector. Second, a modified backprojection process incorporating two weighting factors is defined to address the nonlinearity inherent in conventional backprojection, which otherwise smears each projection along the same trajectory in image space as the original measurement path.  This adaptation leads to a more tractable expression for the PSF. Third, the 3D image-space filtering is computationally intensive and time-consuming, thus, it is generally preferable to perform the filtering in the projection space. However, due to the nonlinear relationship between  a reconstructed pixel and its corresponding projection point on the detector, accurately transferring the filtering operation is challenging. To overcome this issue, we propose an approximate method allows the 3D image-space filtering to be approximately transformed into  2D filtering operations applied to projection space. Under this assumption, a formally simple yet effective 2D directional filter is derived. Finally, a 2D von Hann window and a 2D sinc window are integrated into the filtering process.\\
\indent The remainder of this article is organized as follows. Section 2 begins with the primary knowledge of cone-beam rotary CL. Section 3 presents the derivation of the theorems and the approximate formula for the proposed DFBP method. Experimental results from two sets of actual projection data are provided in Section 4. Finally, Section 5 offers the concluding remarks and a discussion.
\section{Preliminaries of cone-beam rotary CL}
\label{sec:main}
\begin{figure}[!htp]
	\centering
	\includegraphics[width=3.1in]{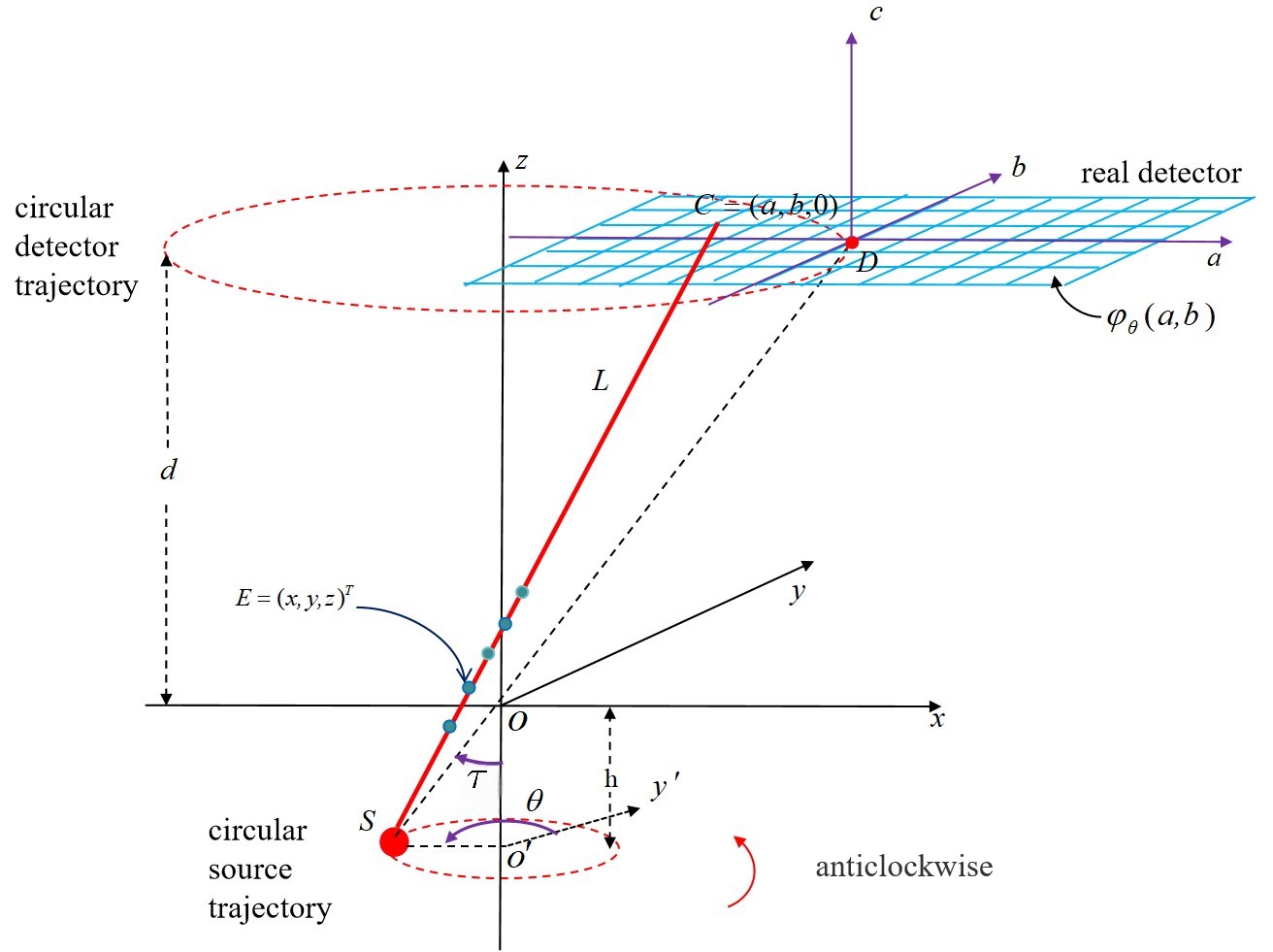}
	\caption{The  cone-beam rotary CL illumination of a flat-panel detector. The X-ray source and detector rotate on a circle with radius $SO\sin\tau $ and $OD\sin\tau $, respectively. The variable $\tau\in (0,90^\circ)$ is the tilted angle.  $y'$ is the projection of $y$-axis on the X-source rotation plane. $\theta$ is the projection angle.   $E$ is a voxels.
The measured projection value at location $C$ along line $L$ is denoted by $\phi_{\theta}(a,b)$.}
	\label{fig1}
\end{figure}

\noindent Cone-beam rotary CL offers a viable solution for  objects with large aspect ratios. This can be achieved by tilting the  object or the source and detector within a conventional CT system, resulting in the X-ray beam intersecting the scanned object at an angle relative to its rotational axis.  In this paper, we focus on the CL scanning modality that involves tilting the source and detector.  \Cref{fig1} illustrates the rotary cone-beam rotary CL with a flat-panel detector.
  According to the geometry illustrated in \Cref{fig1}, in the three-dimensional image coordinate system $\boldsymbol{O\mbox{-}xyz}$, the position of X-ray source 
$\boldsymbol{S}$ is represented by
 \begin{equation}\label{eq:1}
\boldsymbol{S}=\Big(-SO\sin\tau \sin\theta  ,SO\sin\tau \cos\theta  ,-SO\cos\tau \Big),
 \end{equation}
where $SO$ the distance from the rotation center $\boldsymbol{O}$ to the X-ray source $\boldsymbol{S}$.\\
The position of the center of detector 
$\boldsymbol{D}$ is represented by
 \begin{equation}\label{eq:2}
\boldsymbol{D}=\Big(OD\sin\tau \sin\theta  ,-OD\sin\tau \cos\theta  ,OD\cos\tau \Big),
 \end{equation}
where $OD$ the distance from the rotation center $\boldsymbol{O}$ to the detector center $\boldsymbol{D}$.\\
\indent The projection data, denoted by $\phi_{\theta}(a,b)$, is obtained as the line integral of the function $f(x,y,z)$ along the ray path $L$ connecting the source $\boldsymbol{S}$ to the point $\boldsymbol{C}$ on the detector. Here, $\boldsymbol{E}=(x,y,z)$ and $(a,b)$ represents the projection point in the detector coordinate system $\boldsymbol{D\mbox{-}abc}$ where $c=0$. It can be expressed as
 \begin{equation}\label{eq:3}
\phi_{\theta}(a,b)=\int_{-\infty}^{+\infty}f\Big(\boldsymbol{E}+t\frac{\boldsymbol{C}-\boldsymbol{S}}{|\boldsymbol{C}-\boldsymbol{S}|}\Big)dt.
 \end{equation}
\indent In general, the analytic reconstruction of the linear attenuation coefficients $f(x,y,z)$ comprises two principal steps: generating a 3D compound image $p(x,y,z)$ via backprojection the projection $\phi_{\theta}(a,b)$ and subsequently convolving it with an inverse filter function.
\section{The Proposed DFBP Method}
\subsection{Theorem Derivation for DFBP Formulation}
\noindent To derive an accurate expression for \Cref{eq:3}, it is essential to establish  the geometric projection relationship between  a reconstructed pixel  $(x, y, z)^T$  and its corresponding projection point $(a, b)^T$ on the detector. Prior to formulating this relationship, it is necessary to first obtain the coordinate transformation between the detector coordinate system $\boldsymbol{D\mbox{-}abc}$ and the image coordinate system
$\boldsymbol{O\mbox{-}xyz}$.

\begin{lemma}\label{lemma:1}
Let $\tau$ be the tilted angle, $\theta$ the projection angle, and $OD$ the distance from the rotation center $\boldsymbol{O}$ to the detector center $\boldsymbol{D}$. Then,  the  transformation from $\boldsymbol{O\mbox{-}xyz}$  to  $\boldsymbol{D\mbox{-}abc}$ is given by:
\begin{equation}\label{eq:4}
\left(
\begin{array}{ccc}
a \\
b \\
c
\end{array}
\right)
= \left(
\begin{array}{ccc}
x - OD\sin\tau \sin\theta   \\
y + OD\sin\tau \cos\theta   \\
z - OD\cos\tau
\end{array}
\right).
\end{equation}
Conversely, the inverse transformation from $\boldsymbol{D\mbox{-}abc}$ to $\boldsymbol{O\mbox{-}xyz}$ is given by:
\begin{equation}\label{eq:5}
\left(
\begin{array}{ccc}
x \\
y \\
z
\end{array}
\right)
= 
\left(
\begin{array}{ccc}
a + OD\sin\tau\sin\theta \\
b - OD\sin\tau\cos\theta \\
c + OD\cos\tau
\end{array}
\right).
\end{equation}
\end{lemma}
\begin{proof}
 The displacement vector $\overrightarrow{OD}$ in $\boldsymbol{O\mbox{-}xyz}$ coordinates is:
\begin{equation}\label{eq:6}
\overrightarrow{OD} = 
\left(
\begin{array}{ccc}
OD\sin\tau\sin\theta \\
-OD\sin\tau\cos\theta \\
OD\cos\tau
\end{array}
\right).
\end{equation}
Thus, the  transformation from $ \boldsymbol{O\mbox{-}xyz} \to \boldsymbol{D\mbox{-}abc} $, the coordinates in $\boldsymbol{D\mbox{-}abc}$ are obtained by:
 \begin{align}
\left(
\begin{array}{ccc}
a \\
b \\
c
\end{array}
\right)&=
\left(
\begin{array}{ccc}
x \\
y\\
z
\end{array}
\right)-\left(
\begin{array}{ccc}
OD\sin\tau \sin\theta  \\
-OD\sin\tau \cos\theta    \\
OD\cos\tau 
\end{array}
\right)\nonumber\\
&=\left(
\begin{array}{ccc}
x-OD\sin\tau \sin\theta  \\
y+OD\sin\tau \cos\theta    \\
z-OD\cos\tau 
\end{array}
\right).\label{eq:7}
\end{align}
The inverse transformation from $\boldsymbol{D\mbox{-}abc} \to \boldsymbol{O\mbox{-}xyz}$, the original coordinates are recovered by:
 \begin{align}
\left(
\begin{array}{ccc}
x \\
y\\
z
\end{array}
\right)&=
\left(
\begin{array}{ccc}
a \\
b \\
c
\end{array}
\right)+
\left(
\begin{array}{ccc}
OD\sin\tau \sin\theta  \\
-OD\sin\tau \cos\theta    \\
OD\cos\tau 
\end{array}
\right)\nonumber\\
&=\left(
\begin{array}{ccc}
a+OD\sin\tau \sin\theta  \\
b-OD\sin\tau \cos\theta    \\
c+OD\cos\tau 
\end{array}
\right).\label{eq:8}
\end{align}
\end{proof}
\begin{corollary}\label{cor:1}
The coordinates of the X-ray source $\boldsymbol{S}$, defined in the image coordinate system $\boldsymbol{O\mbox{-}xyz}$, in the detector coordinate system $\boldsymbol{D\mbox{-}abc}$ are given by:
\begin{equation}\label{eq:9}
\left(
\begin{array}{ccc}
S_a \\
S_b \\
S_c
\end{array}
\right)
= 
\left(
\begin{array}{ccc}
-SD\sin\tau\sin\theta \\
SD\sin\tau\cos\theta \\
-SD\cos\tau
\end{array}
\right),
\end{equation}
where $SD = SO + OD$ denotes the distance from $S$ to the detector center $D$.
\end{corollary}
\begin{proof}
Substituting \Cref{eq:1} into  \Cref{eq:4} of the \Cref{lemma:1}, we have that
\begin{eqnarray}
\left(
\begin{array}{ccc}
S_a \\
S_b \\
S_c
\end{array}
\right)
&= 
\left(
\begin{array}{ccc}
-SD\sin\tau\sin\theta \\
SD\sin\tau\cos\theta \\
-SD\cos\tau
\end{array}
\right).\label{eq:10}
\end{eqnarray}
\end{proof}

\indent Based on the coordinate transformation in  \Cref{eq:4} from \Cref{lemma:1} and  the  \Cref{eq:9} of the \cref{cor:1}, we establish  the geometric projection relationship between  a reconstructed pixel  $(x, y, z)^T$  and its corresponding projection point $(a, b)^T$ on the detector as follows.
\begin{lemma}\label{lemma:2}
Let $\boldsymbol{E} = (x, y, z)^T$ be a point in the image space and $\boldsymbol{C} = (a, b,0)^T$ be its corresponding projection position in the detector. The following relation then holds:
\begin{equation}\label{eq:11}
a = \frac{SDx\cos\tau -SDz\sin\tau \sin\theta  }{z+SO\cos\tau },
\end{equation}
and
\begin{equation}\label{eq:12}
b = \frac{SDy\cos\tau + SDz\sin\tau \cos\theta  }{z+SO\cos\tau }.
\end{equation}
\end{lemma}
\begin{proof}
As seen from \Cref{fig1}, the source position $\boldsymbol{S}$, the detector point $\boldsymbol{C} = (a,b,0)^{T}$, and the  point $\boldsymbol{E} = (x,y,z)^T$ in the new coordinate system $\boldsymbol{D\mbox{-}abc}$ are collinear.  Using  the coordinate transformation from  \Cref{eq:4} in  \Cref{lemma:1}, we obtain:
\begin{align}\label{eq:13}
\frac{x-OD\sin\tau \sin\theta  - S_{a}}{a - S_{a}} 
&=\frac{z-OD\cos\tau - S_{c}}{0-S_{c}}\\\nonumber
&=\frac{y+OD\sin\tau \cos\theta   - S_{b}}{b - S_{b}}.
\end{align}
From  \Cref{eq:9} in \Cref{cor:1}, it follows that:
\begin{equation}\label{eq:14}
\frac{x+SO\sin\tau \sin\theta  }{a+SD\sin\tau \sin\theta  }=\frac{y-SO\sin\tau \cos\theta  }{b-SD\sin\tau \cos\theta  }=\frac{z+SO\cos\tau }{SD\cos\tau }.
\end{equation}
Therefore, we conclude that
\begin{align}
a &= \frac{SD\cos\tau (x+SO\sin\tau \sin\theta  )}{z+SO\cos\tau }-SD\sin\tau \sin\theta   \nonumber \\
&=\frac{SDx\cos\tau -SDz\sin\tau \sin\theta  }{z+SO\cos\tau },\label{eq:15}
\end{align}
and
\begin{align}
b&=\frac{SD\cos\tau (y-SO\sin\tau \cos\theta  )}{z+SO\cos\tau }+SD\sin\tau \cos\theta  \nonumber \\
&=\frac{SDy\cos\tau + SDz\sin\tau \cos\theta  }{z+SO\cos\tau }.\label{eq:16}
\end{align}

\end{proof}
The point $(x,y,z)$ of the image space is mapped to the  point $(a,b,0)$ on the detector plane via the following nonliner transformation
\begin{equation}\label{eq:17}
\left(
\begin{array}{ccc}
a \\
b
\end{array}
\right)
=H(z)
\left(
\begin{array}{ccc}
\cos\tau  & 0 & -\sin\tau \sin\theta  \\
0  & \cos\tau  & \sin\tau \cos\theta  
\end{array}
\right)
\left(
\begin{array}{ccc}
x \\
y \\
z
\end{array}
\right),
\end{equation}
where $H(z)=\frac{SD}{z+SO\cos\tau }$.

\indent Based on \Cref{eq:5} in \Cref{lemma:1}, together with  \Cref{eq:11} and \Cref{eq:12} in \Cref{lemma:2}, an exact  expression for  \Cref{eq:3} can be derived as follows
\begin{theorem}\label{thm:1}
Under the assumptions of \Cref{lemma:1} and \Cref{lemma:2},  the exact form of  \Cref{eq:3} is given by:

\begin{align}
\phi_{\theta}(a,b) 
&= \int_{-\infty}^{+\infty} 
f\biggl(x + \frac{tSD}{|\boldsymbol{C}-\boldsymbol{S}|}
    \frac{\cos\tau (x+SO\sin\tau \sin\theta)}{z+SO\cos\tau}, \nonumber \\
&\quad y + \frac{tSD}{|\boldsymbol{C}-\boldsymbol{S}|}
    \frac{\cos\tau (y-SO\sin\tau \cos\theta)}{z+SO\cos\tau}, \nonumber \\
&\quad z + \frac{tSD}{|\boldsymbol{C}-\boldsymbol{S}|}\cos\tau \biggr) dt. \label{eq:18}
\end{align}
where \begin{multline}
|\boldsymbol{C}-\boldsymbol{S}| = \\
\sqrt{SD^2 + a^2 + b^2 + 2a\sin\tau \sin\theta - 2bSD\sin\tau \cos\theta},\nonumber
\end{multline}
and
$$ z + SO \cos\tau > 0.$$
\end{theorem}
\begin{proof}
According to  the \Cref{eq:5} of the \Cref{lemma:1}, for a point $\boldsymbol{C}$ with coordinates $(a,b,0)^{T}$ in the coordinate system $\boldsymbol{D\mbox{-}abc}$, its representation in the  coordinate system $\boldsymbol{O\mbox{-}xyz}$ is given by: 
 \begin{equation}\label{eq:19}
\boldsymbol{C}=\left(
\begin{array}{ccc}
a+OD\sin\tau \sin\theta  \\
b-OD\sin\tau \cos\theta  \\
OD\cos\tau 
\end{array}
\right).
 \end{equation}
Using the \Cref{eq:1}, the displacement vector between $\boldsymbol{C}$ and $\boldsymbol{S}$ is given by
 \begin{equation}\label{eq:20}
\boldsymbol{C}-\boldsymbol{S}=\left(
\begin{array}{ccc}
a+SD\sin\tau \sin\theta  \\
b-SD\sin\tau \cos\theta  \\
SD\cos\tau 
\end{array}
\right).
\end{equation}
The magnitude of this vector is then computed as follows:
 \begin{multline}
|\boldsymbol{C}-\boldsymbol{S}| = \\
\sqrt{SD^2 + a^2 + b^2 + 2a\sin\tau \sin\theta - 2bSD\sin\tau \cos\theta}.\label{eq:21}
\end{multline}
Using the  \Cref{eq:20} and \Cref{eq:21}, the exact form of  \Cref{eq:3} is given by:
\begin{align}
\phi_{\theta}(a,b) &=\int_{-\infty}^{+\infty}f\left(x+\frac{t}{|\boldsymbol{C}-\boldsymbol{S}|}(a+SD\sin\tau \sin\theta  ),\right. \nonumber\\
         & y+\frac{t}{|\boldsymbol{C}-\boldsymbol{S}|}(b-SD\sin\tau \cos\theta  ), \nonumber\\
&\left.z+\frac{t}{|\boldsymbol{C}-\boldsymbol{S}|}SD\cos\tau \right)dt. \label{eq:22} 
\end{align}
 Substituting \Cref{eq:11} and (\ref{eq:12}) into Equation (\ref{eq:22}), we obtain:
\begin{align}
\phi_{\theta}(a,b) &=\int_{-\infty}^{+\infty}f\left(x+\frac{tSD}{|\boldsymbol{C}-\boldsymbol{S}|}\frac{\cos\tau (x+SO\sin\tau \sin\theta  )}{z+SO\cos\tau },\right.\nonumber\\
&y+\frac{tSD}{|\boldsymbol{C}-\boldsymbol{S}|}\frac{\cos\tau (y-SO\sin\tau \cos\theta  )}{z+SO\cos\tau }, \nonumber\\
&z+\frac{tSD}{|\boldsymbol{C}-\boldsymbol{S}|}\cos\tau \bigg.\bigg)d t.\label{eq:23} 
\end{align}

In this cone-beam rotary CL geometry, as illustrated in \Cref{fig1}, the object is positioned between the X-ray source and the detector (i.e. $-z<SO  \cos\tau $ if $z<0$). This configuration ensures that $z + SO  \cos\tau  > 0.$
\end{proof}

\indent In traditional backprojection process, the compound image 
$p(x,y,z)$ is generated via a backprojection operation, wherein each measured projection $\phi_{\theta}(a,b)$ is smeared back along the same ray paths in the reconstruction space from which it was originally acquired.  The design of the filter depends on the point spread function, which is obtained by backprojecting the projection of a Dirac delta function $\delta(\boldsymbol{x,y,z})$. Thus, the first step involves computing the projection of a Dirac delta function $\delta(\boldsymbol{x,y,z})$. 
\begin{theorem}\label{thm:2}
Let $\delta(\boldsymbol{x,y,z})$ denotes the Dirac delta function and let $\phi_{\theta}^{\delta}(a, b)$ represent the projection of $\delta(\boldsymbol{x,y,z})$. Then, $\phi_{\theta}^{\delta}(a, b)$ is given by
\begin{multline}\label{eq:27}
\phi_{\theta}^{\delta}(a,b)=\frac{|\boldsymbol{C}-\boldsymbol{S}|}{SD\cos\tau }
\left(\frac{z+SO\cos\tau}{SO\cos\tau}\right)^2 \\
\times\delta\left(x-z\tan\tau \sin\theta\right)
\delta\left(y+z\tan\tau \cos\theta\right).
\end{multline}
\end{theorem}

\begin{proof}
According to  \Cref{eq:18} in \Cref{thm:1},  we have that: 
\begin{align}
\phi_{\theta}^{\delta}(a,b) &=\int_{-\infty}^{+\infty}\delta\left(x+\frac{tSD}{|\boldsymbol{C}-\boldsymbol{S}|}\frac{\cos\tau (x+SO\sin\tau \sin\theta  )}{z+SO\cos\tau },\right.\nonumber\\
&y+\frac{tSD}{|\boldsymbol{C}-\boldsymbol{S}|}\frac{\cos\tau (y-SO\sin\tau \cos\theta  )}{z+SO\cos\tau },\nonumber \\
&z+\frac{tSD}{|\boldsymbol{C}-\boldsymbol{S}|}\cos\tau \bigg.\bigg)d t\nonumber \\
&= \int_{-\infty}^{+\infty} \delta\left(x+\frac{tSD}{|\boldsymbol{C}-\boldsymbol{S}|}\frac{\cos\tau (x+SO\sin\tau \sin\theta  )}{z+SO\cos\tau } \right)\nonumber \\
& \times\delta\left( y+\frac{tSD}{|\boldsymbol{C}-\boldsymbol{S}|}\frac{\cos\tau (y-SO\sin\tau \cos\theta  )}{z+SO\cos\tau }\right)\nonumber \\
&\times\delta\left(z+\frac{tSD}{|\boldsymbol{C}-\boldsymbol{S}|}\cos\tau  \right)dt \nonumber\\
&=\int_{-\infty}^{+\infty} \delta\left(x+\frac{tSD}{|\boldsymbol{C}-\boldsymbol{S}|}\frac{\cos\tau (x+SO\sin\tau \sin\theta  )}{z+SO\cos\tau } \right)\nonumber \\
& \times\delta\left( y+\frac{tSD}{|\boldsymbol{C}-\boldsymbol{S}|}\frac{\cos\tau (y-SO\sin\tau \cos\theta  )}{z+SO\cos\tau }\right)\nonumber \\
&\times\delta\left(\frac{SD\cos\tau }{|\boldsymbol{C}-\boldsymbol{S}|} \left( t + \frac{z|\boldsymbol{C}-\boldsymbol{S}|}{SD\cos\tau }\right)\right) dt.\label{eq:28}
\end{align}

\indent Using the fact  $\delta(kt)=\frac{1}{|k|}\delta(t)$, we have that
\begin{align}
\phi_{\theta}^{\delta}(a,b) 
&= \int_{-\infty}^{+\infty} \delta\left(x+\frac{tSD}{|\boldsymbol{C}-\boldsymbol{S}|}\frac{\cos\tau (x+SO\sin\tau \sin\theta  )}{z+SO\cos\tau } \right)\nonumber \\
&\times\delta\left( y+\frac{tSD}{|\boldsymbol{C}-\boldsymbol{S}|}\frac{\cos\tau (y-SO\sin\tau \cos\theta  )}{z+SO\cos\tau }\right)\nonumber \\
& \times\frac{|\boldsymbol{C}-\boldsymbol{S}|}{SD\cos\tau }\delta\left(t+\frac{z|\boldsymbol{C}-\boldsymbol{S}|}{SD\cos\tau }\right) dt. 
\label{eq:29}
\end{align}
Using the fact  $\int_{-\infty}^{+\infty}f(t)\delta(t-t_0)dt=f(t_0)$, we have that

\begin{align}
\phi_{\theta}^{\delta}(a,b) 
 &=\frac{|\boldsymbol{C}-\boldsymbol{S}|}{SD\cos\tau } \delta\left( \frac{xSO\cos\tau  - zSO\sin\tau \sin\theta  }{z + SO\cos\tau } \right) \nonumber\\
& \times\delta\left( \frac{ySO\cos\tau  + zSO\sin\tau \cos\theta  }{z + SO\cos\tau } \right). \label{eq:30}
\end{align}
The Equation (\ref{eq:30}) can be rewritted as
\begin{align}
\phi_{\theta}^{\delta}(a,b) 
 &= \frac{|\boldsymbol{C}-\boldsymbol{S}|}{SD\cos\tau }\delta\left(\frac{SO\cos\tau }{z+SO\cos\tau }(x-z\tan\tau \sin\theta  )\right)\nonumber \\
&\times\delta\left(\frac{SO\cos\tau }{z+SO\cos\tau }(y+z\tan\tau \cos\theta  )\right).\label{eq:31}
\end{align}

Using the fact  $\delta(kt)=\frac{1}{|k|}\delta(t)$, we have that
\begin{align}
\phi_{\theta}^{\delta}(a,b)
 &= \frac{|\boldsymbol{C}-\boldsymbol{S}|}{SD\cos\tau } \Bigl|\frac{z+SO\cos\tau }{SO\cos\tau }\Bigr|\delta\left(x-z\tan\tau \sin\theta  \right) \nonumber\\
&\times\Bigl| \frac{z+SO\cos\tau }{SO\cos\tau }\Bigr|\delta\left(y+z\tan\tau \cos\theta  \right).\label{eq:32}
\end{align}
Since $z + SO  \cos\tau  > 0 $ in \Cref{thm:1},  it follows that
\begin{multline}\label{eq:33}
\phi_{\theta}^{\delta}(a,b)=\frac{|\boldsymbol{C}-\boldsymbol{S}|}{SD\cos\tau }
\left(\frac{z+SO\cos\tau}{SO\cos\tau}\right)^2 \\
\times\delta\left(x-z\tan\tau \sin\theta\right)
\delta\left(y+z\tan\tau \cos\theta\right).
\end{multline}
\end{proof}

To obtain the point spread function, the  projection data $\phi_{\theta}^{\delta}(a,b)$ is backprojected into the image space. This is achieved by smearing its values along the same ray paths from which the projection data was originally acquired.
  However, in cone-beam rotary CL, the presence of the weighting factor
$$\frac{|\boldsymbol{C}-\boldsymbol{S}|}{SD\cos\tau }(\frac{z+SO\cos\tau }{SO\cos\tau })^2,$$
introduces a nonlinear dependence on the spatial coordinates  $z$. This nonlinearity  makes it ineffective to simply smear back each measured projection  $\phi_{\theta}^{\delta}(a,b)$  along the original ray paths. This not only complicates the design of the filter but also renders the approach impractical for  implementation.\\
 \indent To overcome this challenge, two weighting factor are  introduced into the backprojection process. The first weight is applied in projection space with weight  $\frac{SO\cos\tau }{|\boldsymbol{C}-\boldsymbol{S}|}$. The weighted projection is then smeared back along the same ray paths in the image space from which it was originally acquired. This process is expressed as
 \begin{equation}\label{eq:34}
 p_{\theta}^{\delta,w}(x,y,z) =\phi_{\theta}^{\delta}(a,b)\frac{SO\cos\tau }{|\boldsymbol{C}-\boldsymbol{S}|}.
 \end{equation} 
The second weight is applied in image space with weight  $\frac{(SO\cos\tau )^2}{(z+SO\cos\tau )^2}$, 
\begin{equation}\label{eq:35}
p_{\theta}^{\delta}(x,y,z)=\frac{(SO\cos\tau )^2}{(z+SO\cos\tau )^2}p_{\theta}^{\delta,w}(x,y,z).
 \end{equation}
 
\indent As will be shown later, the introduction of these two weighting factors significantly simplifies the expression of the filter. Therefore, we define a new backprojection process for cone-beam rotary CL as follows:
\begin{definition}\label{def:1}
Given the projection data $\phi_{\theta}(a,b)$, the backprojection process for cone-beam rotary CL is defined as
 \begin{equation}\label{eq:36}
p(x,y,z) =\frac{1}{2\pi}\int_{0}^{2\pi}\frac{(SO\cos\tau )^2}{(z+SO\cos\tau )^2}\phi_{\theta}^{w}(a,b) d\theta.
 \end{equation} 
where \begin{multline}
\phi_{\theta}^{w}(a,b)=\\
\frac{SO\cos\tau \phi_{\theta}(a,b)}{
\sqrt{SD^2+a^2+b^2+2aSD\sin\tau \sin\theta -2bSD\sin\tau \cos\theta}}.\nonumber
\end{multline}
\end{definition}
Now, we will give the PSF of the projection and back-projection procesess.
\begin{theorem}\label{thm:3}
 Let $\delta(x,y,z)$ denotes a  Dirac delta function, and let  $\phi_{\theta}^{\delta}(a,b)$ be the projection of $\delta(x,y,z)$. The back-projection process is defined as \Cref{eq:36}. Then, the point spread function of the combined projection and back-projection process is given by
 \begin{equation}\label{eq:37}
g(x,y,z)= \left\{ \begin{array}{cc}  
\frac{1}{2\pi|z\tan\tau |}\frac{SO}{SD}, & if\;\; x^2+y^2=(z\tan\tau )^2,\\
0, & otherwise.  
\end{array}\right.  
 \end{equation} 
\end{theorem}

\begin{proof}
\indent According to \Cref{eq:36} of \Cref{def:1}, the point spread function $g(x,y,z) $ can be written as
\begin{equation}\label{eq:38}
g(x,y,z) = \frac{1}{2\pi} \int_{0}^{2\pi}  \frac{(SO\cos\tau )^2}{(z+SO\cos\tau )^2}\phi_{\theta}^{w,\delta}(a,b) \, d\theta.
\end{equation}
where $\phi_{\theta}^{w,\delta}(a,b) =\frac{SO\cos\tau \phi_{\theta}^{\delta}(a,b)}{|\boldsymbol{C}-\boldsymbol{S}|}.$\\
\indent According to Equation (\ref{eq:27}) in \Cref{thm:2}, \Cref{eq:38} can be simplified as follows: 
\begin{multline}\label{eq:39}
g(x,y,z) = \frac{1}{2\pi}\frac{SO}{SD} \\
\times\int_{0}^{2\pi} \delta\left( x -z\tan\tau \sin\theta \right)
\delta\left(y+z\tan\tau \cos\theta \right) d\theta.
\end{multline}
\indent For the transformation:
\begin{equation}\label{eq:40}
u = x - z\tan\tau \sin\theta  , \quad v = y + z\tan\tau \cos\theta  .
\end{equation}
The elements of the Jacobian matrix are:
\begin{equation}\label{eq:41}
\frac{\partial u}{\partial \theta} = -z\tan\tau \cos\theta  , \quad \frac{\partial v}{\partial \theta} = -z\tan\tau \sin\theta  .
\end{equation}
And the Jacobian determinant is:
\begin{equation}\label{eq:42}
J = \sqrt{(\frac{\partial u}{\partial \theta})^2+(\frac{\partial v}{\partial \theta})^2}=|z\tan\tau |.
\end{equation}

\indent According to \Cref{eq:40}, \Cref{eq:41}  and \Cref{eq:42}, a variable substitution is applied to the integral Equation (\ref{eq:39}).
\begin{equation}\label{eq:43}
g(x,y,z)=\frac{1}{2\pi}\frac{SO}{SD}\int_{-\infty}^{\infty} \int_{-\infty}^{\infty} \delta(u)\delta(v) \frac{1}{|z\tan\tau |} du dv.
\end{equation}

\indent When $x^2 + y^2 = z^2 \tan^2\tau $, i.e., $u=0$ and $v=0$, which leads to 
\begin{equation}\label{eq:44}
g(x,y,z) = \frac{1}{2\pi|z\tan\tau |}\frac{SO}{SD}.
\end{equation}

\indent When $x^2 + y^2 \neq z^2 \tan^2\tau $, i.e., $u\neq 0$ and $v\neq 0$, which leads to 
\begin{equation}\label{eq:45}
g(x,y,z)=0.
\end{equation}
Thus, the point function is given as follows
 \begin{equation}\label{eq:46}
g(x,y,z)= \left\{ \begin{array}{cc}  
\frac{1}{2\pi |z\tan\tau |}\frac{SO}{SD}, & if\;\; x^2+y^2=(z\tan\tau )^2,\\
0, & otherwise.  
\end{array}\right.
 \end{equation} 
\end{proof}

After obtaining the point spread function $g(x,y,z)$, the back-projection image $p(x,y,z)$ is expressed using $f(x,y,z)$ and $g(x,y,z)$ as
 \begin{equation}\label{eq:47}
p(x,y,z) =f(x,y,z)\ast g(x,y,z),
 \end{equation} 
where $\ast$ denotes three-dimensional convolution.\\
\indent According to the convolution theorem, we have that
 \begin{equation}\label{eq:48}
P(\omega_x, \omega_y,\omega_z) =F(\omega_x, \omega_y,\omega_z) G(\omega_x, \omega_y,\omega_z).
 \end{equation} 
where $P(\omega_x, \omega_y, \omega_z)$, $F(\omega_x, \omega_y, \omega_z)$ and  $ G(\omega_x, \omega_y,\omega_z)$ are the three-dimensional Fourier transforms of the backprojection image  $p(x, y, z)$, the original image $f(x, y, z)$  and the point spread function $g(x, y, z)$, respectively.\\
\indent According to \Cref{eq:48}, the three-dimensional Fourier transform of $f(x, y, z)$ can be obtained as 
 \begin{equation}\label{eq:49}
F(\omega_x, \omega_y,\omega_z)=\frac{P(\omega_x, \omega_y,\omega_z) }{G(\omega_x, \omega_y,\omega_z)}.
 \end{equation} 
Consequently, when the point spread function $g(x,y,z)$ is known, the attenuation coefficient   $f(x, y, z)$  can be reconstucted by 
\begin{equation}\label{eq:50}
\begin{split}
f(x,y,z) &= \mathcal{F}^{-1}\left( \frac{P(\omega_x, \omega_y,\omega_z)}{G(\omega_x, \omega_y,\omega_z)} \right) \\
&= \mathcal{F}^{-1}\left( P(\omega_x, \omega_y,\omega_z)G_{\text{inv}}(\omega_x, \omega_y,\omega_z) \right).
\end{split}
\end{equation}
\indent In \Cref{eq:50}, $P(\omega_x, \omega_y, \omega_z)$ is obtained by applying a three-dimensional Fourier transform to the back-projection image $p(x, y, z)$, which is  given by \Cref{eq:36} in \Cref{def:1}. In the following, we will derive an explicit expression for the inverse filter $G_{\rm{inv}}(\omega_x, \omega_y,\omega_z)$.
\begin{theorem}\label{thm:4}
Let $\tau\in (0,90^{\circ})$,  the Fourier transform of the point spread function $g(x,y,z)$  is expressed  in   cylindrical coordinates as 
\begin{equation}\label{eq:51}
G (\boldsymbol{\omega_{Cyl}})=
\begin{cases}
\dfrac{2SO}{SD \cdot H(\omega_k, \omega_z, \tau)}, & if\;|\omega_z|<\omega_{k}\tan\tau, \\
0, & \text{otherwise},
\end{cases}
\end{equation}
where  $H(\omega_k, \omega_z, \tau)=\tan\tau \sqrt{\omega_{k}^{2} - \omega_z^2\cot^2\tau }$  and $\boldsymbol{\omega_{Cyl}}=(\omega_{k},\omega_{\phi},  \omega_z)$. \\
\indent The multiplicative inverse of $G(\omega_{k},\omega_{\phi},  \omega_z)$ is
 \begin{equation}\label{eq:52}
G_{inv}(\boldsymbol{\omega_{Cyl}})  =
\left\{ \begin{array}{ll}
\dfrac{SD \cdot H(\omega_k, \omega_z, \tau)}{ 2SO},  & if\;  |\omega_z|\leq\omega_{k}\tan\tau , \\
0, & otherwise.
\end{array} \right.
 \end{equation} 

\indent The multiplicative inverse of  $G(\omega_{k},\omega_{\phi},  \omega_z)$ is expressed in  Cartesian  coordinates as 
\begin{equation}\label{eq:53}
G_{\text{inv}}(\boldsymbol{\omega})=
\begin{cases}
\dfrac{SD \tan\tau}{2SO} \sqrt{\omega_{k}^{2} - \omega_z^2\cot^2\tau}, & 
\text{if\; } |\omega_z| \leq \omega_{k} \tan\tau, \\
0, & \text{otherwise},
\end{cases}
\end{equation}
where $\boldsymbol{\omega} = (\omega_x, \omega_y, \omega_z)$ and $\omega_{k} = \sqrt{\omega_x^2 + \omega_y^2}$.
\end{theorem}

\begin{proof}
According to \Cref{eq:37} of \Cref{thm:3},  the point spread function  is
 \begin{equation}\label{eq:54}
g(x,y,z)=\left\{ \begin{array}{cc}  
\frac{1}{2\pi} \frac{1}{|z\tan\tau |}\frac{SO}{SD}, & if\;\; x^2+y^2=(z\tan\tau )^2,\\
0, & otherwise.  
\end{array}\right. 
 \end{equation}
It can be rewritten as
 \begin{equation}\label{eq:55}
g(x,y,z) = \frac{1}{2\pi|z \tan\tau |}  \frac{SO}{SD}\delta(r - |z \tan\tau |),
\end{equation}
where $r=\sqrt{x^2+y^2}$ and $\delta(\cdot)$ is the Dirac delta function enforcing the circular support at radius $|z \tan\tau |$.\\
 \indent Usig the cylindrical coordinates
 \begin{equation}\label{eq:56}
\left\{ \begin{array}{ll} 
\omega_{k}=\sqrt{\omega_{x}^2+\omega_{y}^2}, \\
\omega_{\phi} = \arctan(\omega_x / \omega_y),\\
\omega_z=\omega_z.
\end{array} \right.
\end{equation}
The 2D Fourier transform of  the point spread function $g(x,y,z)$  becomes:
\begin{multline}\label{eq:57}
\mathcal{F}\{g\}(\omega_{k}, \omega_{\phi}, z) 
= \frac{1}{2\pi R}\frac{SO}{SD} \\
\times\int_0^{2\pi} \int_0^\infty \delta(r - R)  
e^{-i \omega_{k} r q(\theta) } \, r \, dr \, d\theta,
\end{multline}
where $R = |z \tan\tau|$, $\omega_{k} = \sqrt{\omega_x^2 + \omega_y^2}$ and $q(\theta)  =\cos(\theta - \omega_{\phi})$ and $\omega_{\phi} = \arctan(\omega_x /\omega_y)$.

Using the fact  $\int_{-\infty}^{+\infty}f(t)\delta(t-t_0)dt=f(t_0)$, we have that
\begin{equation}\label{eq:58}
\mathcal{F}\{g\}(\omega_{k},\omega_{\phi}, z)  =\frac{1}{2\pi} \frac{SO}{SD} \int_0^{2\pi} e^{-i \omega_{k} |z \tan\tau | \cos(\theta - \omega_\phi)} \, d\theta.
\end{equation}
\indent The angular integral is a representation of the Bessel function $ J_0$, which is derived by employing a variable substitution and exploiting the periodicity of the integrand.
\begin{align}
\int_{0}^{2\pi} e^{-i \omega_{k} |z \tan\tau| \cos(\theta - \omega_\phi)} \, d\theta
&= \int_{-\omega_\phi}^{2\pi - \omega_\phi} e^{-i \omega_{k} |z \tan\tau| \cos\xi} \, d\xi \nonumber \\
&= \int_{0}^{2\pi} e^{-i \omega_{k} |z \tan\tau| \cos\xi} \, d\xi \nonumber \\
&= 2\pi J_{0}(\omega_{k} |z \tan\tau|). \label{eq:59}
\end{align}
\indent Since $\tau \in (0, 90^\circ)$, we have $\tan(\tau) > 0$. The Fourier transform of the Bessel function $J_0\left( \omega_k |z \tan\tau | \right)$ is given by:
\begin{equation}\label{eq:61}
\mathcal{F}\{ J_0\left( \omega_{k} |z \tan\tau |\right) \}(\omega_z) = \int_{-\infty}^{\infty}  J_0\left(  \omega_{k}\tan\tau |z |\right)  e^{-i \omega_z z} \, dz.
\end{equation}
\Cref{eq:61} can be converted into the following form
\begin{align}
\mathcal{F}\{J_0(\omega_k |z\tan\tau|)\}(\omega_z)
&= \int_{-\infty}^{\infty} J_0(\omega_k \tan\tau |z|) \cos(\omega_z z) \, dz \nonumber \\
& - i \int_{-\infty}^{\infty} J_0(\omega_k \tan\tau |z|) \sin(\omega_z z) \, dz. \label{eq:62}
\end{align}
Since $J_0\left(\omega_{k}\tan\tau  |z  |\right) $ is even, the term $J_0\left( \omega_{k}\tan\tau  |z |\right)\sin( \omega_z z)$ is odd, and its integral over $(-\infty,\infty)$ vanishes, the term $J_0\left( \omega_{k}\tan\tau|z|\right)\cos( \omega_z z)$ is even, the transform simplifies to:
\begin{equation}\label{eq:63}
\mathcal{F}\{J_0\left( \omega_{k} |z \tan\tau |\right)\}(\omega_z) = 2 \int_{0}^{\infty} J_0\left(  \omega_{k}\tan\tau z\right) \cos(\omega_z z) \, dz.
\end{equation}
From integral tables \cite{gradshteyn2014table,matsuo1993three}:
\begin{equation}\label{eq:64}
\int_{0}^{\infty} J_0(A t) \cos(Bt) \, dt = 
\left\{ \begin{array}{ll} 
\frac{1}{\sqrt{A^2 - B^2}}, & if\; |B|< A, \\
0, &  otherwise.
\end{array} \right.
\end{equation}
 Thus, the Fourier transform of $ J_0\left( \omega_{k} |z \tan\tau |\right) $ is:
\begin{multline}\label{eq:65}
\int_{-\infty}^{\infty}  
J_0\!\bigl( \omega_k |z \tan\tau| \bigr) 
e^{-\mathrm{j} \omega_z z} \, dz 
=\\ \begin{cases}
\dfrac{2}{\sqrt{(\omega_k \tan\tau)^2 - \omega_z^2}}, 
& |\omega_z| < \omega_k \tan\tau, \\[8pt]
0, & otherwise.
\end{cases}
\end{multline}
Thus, the Fourier transform of the point spread function $g(x,y,z)$  is expressed  in   cylindrical coordinates as 
\begin{align}
G(\omega_{k},\omega_{\phi},  \omega_z)& = \frac{SO}{SD}\int_{-\infty}^{\infty} J_0\left( \omega_{k} |z \tan\tau | \right) e^{-i \omega_z z} dz\nonumber\\
&=\left\{ \begin{array}{ll} 
\frac{SO}{SD}\frac{2}{\sqrt{(\omega_{k}\tan\tau )^2 - \omega_z^2}},& if \: |\omega_z| < \omega_{k}\tan\tau , \\
0, & otherwise.
\end{array} \right.\nonumber\\
&= \left\{ \begin{array}{ll} 
\dfrac{2SO}{SD \cdot H(\omega_k, \omega_z, \tau)}, & if \: |\omega_z| < \omega_{k}\tan\tau , \\
0,  &otherwise.\label{eq:66}
\end{array} \right.
\end{align}
where  $H(\omega_k, \omega_z, \tau)=\tan\tau \sqrt{\omega_{k}^{2} - \omega_z^2\cot^2\tau }$. \\
\indent The multiplicative inverse of $G(\omega_{k},\omega_{\phi},  \omega_z)$ is
\begin{equation}
G_{inv}(\boldsymbol{\omega_{Cyl}})  =
\left\{ \begin{array}{ll}
\dfrac{SD \cdot H(\omega_k, \omega_z, \tau)}{ 2SO},  & if\;  |\omega_z|\leq\omega_{k}\tan\tau , \\
0, & otherwise.
\end{array} \right.
\end{equation}
where  $\boldsymbol{\omega_{Cyl}}=(\omega_{k},\omega_{\phi},  \omega_z)$.\\
\indent According \Cref{eq:56}, the multiplicative inverse of  $G(\omega_{k},\omega_{\phi},  \omega_z)$ is expressed in  Cartesian  coordinates as 
 \begin{equation}\label{eq:68}
G_{\text{inv}}(\boldsymbol{\omega})=
\begin{cases}
\dfrac{SD \tan\tau}{2SO} \sqrt{\omega_{k}^{2} - \omega_z^2\cot^2\tau}, & 
\text{if\; } |\omega_z| \leq \omega_{k} \tan\tau, \\
0, & \text{otherwise},
\end{cases}
\end{equation}
where $\boldsymbol{\omega} = (\omega_x, \omega_y, \omega_z)$ and $\omega_{k} = \sqrt{\omega_x^2 + \omega_y^2}$.
\end{proof}

\indent
Performing filtering operations directly in the image space is computationally intensive and time-consuming. One strategy to mitigate this issue is to shift the filtering process into the projection space. In parallel-beam CT, the central slice  theorem can be employed to convert image space filtering into operations on projection space. However, this theorem is not applicable to cone-beam rotary CL due to its distinct geometric configuration.  The relationship between a reconstructed pixel  $(x,y,z)$  and its corresponding projection point  $(a,b)$ on the detector is nonlinear, making it challenging to accurately transfer the filtering operation from the image space to the projection space. To address this limitation, we propose an approximate method that avoids direct computation of the 3D Fourier transform. Instead, we first compute the 2D Fourier transform with respect to $x$ and $y$, and then introduce an assumption based on the cone-beam rotary CL geometry to linearize the relationship between  $(a,b)$ and  $(x,y)$. This enables the subsequent computation of a 1D Fourier transform with respect to $z$. Through this approach, the image-space filtering can be approximately transformed into operations on projection data, and it was given in the following theorem.
\begin{theorem}\label{thm:5}
 Let $p_{\theta}(x,y,z)= \frac{(SO\cos\tau )^2}{(z+SO\cos\tau )^2} \phi_{\theta}^{w}(a,b) $, under the assume that   $|z| \ll SO\cos\tau $, the Fourier transform of  $p_{\theta}(x,y,z)$  is given by
 \begin{multline}\label{eq:thm51}
\mathcal{F}(p_{\theta})(\omega_x, \omega_y, \omega_z)    \\
 \approx\left\{ \begin{array}{ll}
2\pi(\frac{SO}{SD})^2 \mathcal{F}(\phi_{\theta}^{w})(\omega_{a}, \omega_{b}),& if \;\omega_{xyz}=0,\\
0, & otherwise.
\end{array} \right.
 \end{multline} 
where   $\omega_{xyz}=\sin\tau (\omega_x\sin\theta  -\omega_y\cos\theta  )+\omega_z\cos\tau $. \\
\indent The multiplicative inverse of $G(\omega_{x},\omega_{y},  \omega_{z})$  in \Cref{thm:4}  can be approximated as
 \begin{equation}\label{eq:thm52}
G_{inv}(\omega_{x},\omega_{y},  \omega_{z}) \approx 
(\frac{SD}{SO})^2\frac{\tan\tau |\omega_{a}\cos\theta  +\omega_{b}\sin\theta  |}{2}.
\end{equation}
\indent This allows the transformation of image space filtering into operations on projection data:
 \begin{equation}\label{eq:thm53}
\begin{aligned}
\mathcal{F}(p_{\theta})(\omega_{x}, \omega_{y}, \omega_{z}) 
&G_{\text{inv}}(\omega_{x}, \omega_{y}, \omega_{z}) \\
&\approx 2\pi \mathcal{F}(\phi_{\theta}^{w})(\omega_{a}, \omega_{b}) 
H_{\text{dir}}(\omega_{a}, \omega_{b}).
\end{aligned}
\end{equation}
where $H_{dir}(\omega_{a},\omega_{b})=\frac{\tan\tau |\omega_{a}\cos\theta  +\omega_{b}\sin\theta  |}{2}$.
\end{theorem}
\begin{proof}
Let  \begin{equation}\label{eq:69}
p_{\theta}(x,y,z)=\frac{(SO\cos\tau )^2}{(z+SO\cos\tau )^2}\phi_{\theta}^{w}(a,b).
 \end{equation}
where  \begin{multline}
\phi_{\theta}^{w}(a,b)=\\
 \frac{SO\cos\tau \phi_{\theta}(a,b)}{\sqrt{SD^2+a^2+b^2+2aSD\sin\tau \sin\theta  -2bSD\sin\tau \cos\theta  }}.\nonumber \end{multline}\\
\indent We begin with the two-dimensional Fourier transform:
\begin{align}
\mathcal{F}(p_{\theta})(\omega_x, \omega_y, z) &= \int_{-\infty}^{\infty}\int_{-\infty}^{\infty} p_{\theta}(x,y,z) e^{-i (\omega_x x + \omega_y y)} dx dy \nonumber\\
&=\int_{-\infty}^{\infty}\int_{-\infty}^{\infty}\frac{(SO\cos\tau )^2\phi_{\theta}^{w}(a,b)}{(z+SO\cos\tau )^2}\nonumber\\ &\quad\quad\quad\quad\quad e^{-i (\omega_x x + \omega_y y )} dx dy.\label{eq:70}
\end{align}

According to \Cref{eq:11} and \Cref{eq:12}, we have that
 \begin{equation}\label{eq:71}
 x=\frac{a(z+SO\cos\tau )+SDz\sin\tau \sin\theta  )}{SD\cos\tau },
 \end{equation}
and
 \begin{equation}\label{eq:72}
y=\frac{b(z+SO\cos\tau )-SDz\sin\tau \cos\theta  )}{SD\cos\tau }.
 \end{equation}
The Jacobian determinant of the transformation $(x,y) \to (a,b)$ is:
\begin{equation}\label{eq:76}
 dxdy=|J|dadb,
 \end{equation}
where
\begin{align}\label{eq:77}
J &=\mathrm{det}\left(\frac{\partial(x,y)}{\partial(a,b)}\right)\nonumber\\
&=\left| \begin{array}{c c }  
\frac{\partial x}{\partial a} &   \frac{\partial x}{\partial b}\\
\frac{\partial y}{\partial a} &   \frac{\partial y}{\partial b}\\
\end{array}\right|\nonumber \\
&=
\left| \begin{array}{c  c } 
\frac{z+SO\cos\tau }{SD\cos\tau } &0\\
0&  \frac{z+SO\cos\tau }{SD\cos\tau }\\
\end{array}\right|\nonumber\\
&= (\frac{z+SO\cos\tau }{SD\cos\tau })^2.
\end{align}
Now consider the phase term:
 \begin{align}
\omega_x x + \omega_y y &=\omega_x(\frac{a(z+SO\cos\tau )+SDz\sin\tau \sin\theta  }{SD\cos\tau })\nonumber\\
&+\omega_y (\frac{b(z+SO\cos\tau )-SDz\sin\tau \cos\theta  }{SD\cos\tau }).\label{eq:98}
\end{align}

By rearranging the Equation (\ref{eq:98}), we obtain
 \begin{equation}\label{eq:79}
\omega_x x + \omega_y y =\omega_a a+\omega_b b+\frac{z\sin\tau (\omega_x\sin\theta  -\omega_y\cos\theta  )}{\cos\tau }.
\end{equation}

where
$\omega_{a}=\omega_x\frac{z+SO\cos\tau }{SD\cos\tau }$ and $\omega_{b}=\omega_y\frac{z+SO\cos\tau }{SD\cos\tau }.$

The Fourier transform then becomes:
 \begin{align}\label{eq:80}
\mathcal{F}(p_{\theta})(\omega_x, \omega_y, z) &= M\int_{-\infty}^{\infty}\int_{-\infty}^{\infty} \frac{SO^2}{SD^2}\nonumber\\
& \quad\quad\quad\quad\times\phi_{\theta}^{w}\left( a,b \right) e^{-i (\omega_{a} a + \omega_{b} b )} da  db \nonumber\\
&=(\frac{SO}{SD})^2M\mathcal{F}(\phi_{\theta}^{w})(\omega_{a}, \omega_{b}),
 \end{align}
where $M=e^{-i \frac{z\sin\tau (\omega_x\sin\theta  -\omega_y\cos\theta  )}{\cos\tau }}$.\\
\indent For cone-beam rotary CL, the object is a thin plate-like structure, see \Cref{fig1}, which confines our analysis to  $|z| \ll SO\cos\tau $,
 \begin{equation}\label{eq:81}
\omega_{a}\approx \omega_x\frac{SO}{SD}, \quad \omega_{b}\approx\omega_y\frac{SO}{SD}.
\end{equation}
 The Fourier transform on $z$ then becomes:
\begin{align}
\mathcal{F}(p_{\theta})(\omega_x, \omega_y, \omega_z) 
&= \int_{-\infty}^{\infty} \left(\frac{SO}{SD}\right)^2 
   e^{-i \frac{z\sin\tau (\omega_x\sin\theta - \omega_y\cos\theta)}{\cos\tau}} \nonumber\\
&\quad\quad\quad\quad\times   \mathcal{F}(\phi_{\theta}^{w})(\omega_{a}, \omega_{b})   e^{-i\omega_{z}z} \, dz \nonumber \\
&\approx \left(\frac{SO}{SD}\right)^2 
   \mathcal{F}(\phi_{\theta}^{w})(\omega_{a}, \omega_{b}) \nonumber\\
&\quad\times    \int_{-\infty}^{\infty} 
   e^{-i \left(\frac{\sin\tau (\omega_x\sin\theta - \omega_y\cos\theta)}{\cos\tau} + \omega_{z}\right)z} 
   \, dz \nonumber \\
&\approx 2\pi \left(\frac{SO}{SD}\right)^2 
   \mathcal{F}(\phi_{\theta}^{w})(\omega_{a}, \omega_{b})
   \delta(\omega_{xyz}). \label{eq:83}
\end{align}

where $\omega_{xyz}=\sin\tau (\omega_x\sin\theta  -\omega_y\cos\theta  )+\omega_z\cos\tau $.\\
\indent  We have:
 \begin{multline}\label{eq:85}
\mathcal{F}(p_{\theta})(\omega_x, \omega_y, \omega_z)   \\ 
 \approx\left\{ \begin{array}{ll}
 2\pi(\frac{SO}{SD})^2 \mathcal{F}(\phi_{\theta}^{w})(\omega_{a}, \omega_{b}),& if \;\omega_{xyz}=0,\\
0, & otherwise.
\end{array} \right.
 \end{multline}
According to \Cref{eq:81} and $\omega_{xyz}=0$,  we have that
 \begin{equation}\label{eq:86}
 \left\{ \begin{array}{ll}
\omega_x\approx\frac{SD}{SO}\omega_a,\\
\omega_y\approx \frac{SD}{SO}\omega_b,\\
\omega_z=\frac{SD}{SO}\frac{\omega_b\cos\theta  \sin\tau -\omega_a\sin\theta  \sin\tau }{\cos\tau }.
\end{array} \right.
\end{equation}
Thus, we have that
 \begin{equation}\label{eq:88}
\omega_x^2 + \omega_y^2-\omega_z^2\cot^2\tau \approx(\frac{SD}{SO})^2(\omega_{a}\cos\theta  +\omega_{b}\sin\theta  )^{2}.
\end{equation}

\indent Submiting \Cref{eq:88} into Equation (\ref{eq:53}) in \Cref{thm:4}, the multiplicative inverse $G_{inv}(\omega_x, \omega_y, \omega_z)$  is approximated as
 \begin{multline}\label{eq:89}
G_{inv}(\omega_{x},\omega_{y},  \omega_{z})\\
\approx \left\{ \begin{array}{ll}
(\frac{SD}{SO})^2\frac{\tan\tau |\omega_{a}\cos\theta  +\omega_{b}\sin\theta  |}{2},&  if \; |\omega_z|\leq \omega_{k}\tan\tau , \\
0, & otherwise.
\end{array} \right.
 \end{multline}
where  $\omega_{k}=\sqrt{\omega_{x}^2+\omega_{y}^2}$.\\
\indent According to \Cref{eq:88}, the condition $|\omega_z|\leq\sqrt{\omega_{x}^2+\omega_{y}^2} \tan\tau $ always holds  approximately. This can be verified by the following derivation:
 \begin{align}\label{eq:89x}
\omega_{x}^2+\omega_{y}^2-\frac{|\omega_z|^2}{\tan^{2}\tau }&=\omega_x^2 + \omega_y^2-\omega_z^2\cot^2\tau \nonumber\\
&\approx(\frac{SD}{SO})^2(\omega_{a}\cos\theta  +\omega_{b}\sin\theta  )^{2}\geq 0.
 \end{align}
Therefore,
 \begin{equation}\label{eq:91}
G_{inv}(\omega_{x},\omega_{y},  \omega_{z}) \approx 
(\frac{SD}{SO})^2\frac{\tan\tau |\omega_{a}\cos\theta  +\omega_{b}\sin\theta  |}{2}.
\end{equation}
Finally, the product becomes:
\begin{equation}\label{eq:92}
\begin{aligned}
\mathcal{F}(p_{\theta})(\omega_{x}, \omega_{y}, \omega_{z}) 
&G_{\text{inv}}(\omega_{x}, \omega_{y}, \omega_{z}) \\
&\approx 2\pi \mathcal{F}(\phi_{\theta}^{w})(\omega_{a}, \omega_{b}) 
H_{\text{dir}}(\omega_{a}, \omega_{b}).
\end{aligned}
\end{equation}
where $H_{dir}(\omega_{a},\omega_{b})=\frac{\tan\tau |\omega_{a}\cos\theta  +\omega_{b}\sin\theta  |}{2}$.
\end{proof}
\subsection{Proposed  Formula}
\indent As established by \Cref{thm:5}, 3D filtering in image space can be approximately substituted by computationally more efficient 2D directional filtering, where $\theta$ denotes the filtering direction. This 2D directional filtering approach differs from traditional FBP or FDK methods, which employ 1D filtering along detector rows. It should be noted that in the proposed method, artifacts tend to increase in reconstructed slices located farther from the central plane due to the approximate condition $|z| \ll SO\cos\tau $. Nevertheless, the proposed method consistently demonstrates superior artifact suppression compared to the classical FBP or FDK techniques. \\
\indent To suppress noise in the projection data, a 2D von Hann window is incorporated into the filtering process due to its favorable spectral characteristics.
In the projection space, the 2D von Hann window is expressed as follows:
\begin{multline}\label{eq:93}
H_{hann}(\omega_a,\omega_b)=\\
 \left\{ \begin{array}{ll}
\frac{\left(1 + \cos\left(\frac{\omega_a}{B_{u}}\right)\right)}{2\pi B_{u}}\left(1 + \cos\left(\frac{\omega_b}{B_{u}}\right)\right), & if  \sqrt{\omega_{a}^2+\omega_{b}^2} < \pi B_{u},\\
0,  & otherwise.
\end{array} \right.
\end{multline}
where $B_{u}$ represents an adjustable parameter governing the spectral cutoff.
\indent Additionally, to improve the signal to noise ratio, a 2D $sinc$ window is incorporated into the filtering process:
\begin{multline}\label{eq:94}
H_{sinc}(\omega_{a},\omega_{b})\\
= \left\{ \begin{array}{ll}
sinc\left(\frac{\sqrt{\omega_{a}^2+\omega_{b}^2}}{B_{v}}\right), & if \; \sqrt{\omega_{a}^2+\omega_{b}^2} < \pi B_{v}, \\
0, & otherwise.
\end{array} \right.
\end{multline}
where $B_{v}$ represents another adjustable parameter governing the spectral cutoff.\\
\indent Finally, according to \Cref{eq:36}, \Cref{eq:50}, \Cref{eq:thm53}, \Cref{eq:93} and \Cref{eq:94},  the approximate directional filtering backprojection  method (DFBP) for cone-Beam Rotary Computed Laminography can be expressed as follows
\begin{equation}\label{fbp99}
f(x,y,z)\approx  \int_{0}^{2\pi}\mathcal{F}^{-1}\bigg(\mathcal{F}(\phi_{\theta}^{w})H_{dir}H_{hann}H_{sinc}(\omega_{a},\omega_{b})\bigg) d\theta.
\end{equation}
where 
\begin{multline}
\phi_{\theta}^{w}(a,b)=\nonumber\\
\frac{SO\cos\tau \phi_{\theta}(a,b)}{\sqrt{SD^2+a^2+b^2+2aSD\sin\tau \sin\theta  -2bSD\sin\tau \cos\theta  }},
\end{multline}
$$
a = \frac{SDx\cos\tau -SDz\sin\tau \sin\theta  }{z+SO\cos\tau },
$$
$$
b = \frac{SDy\cos\tau + SDz\sin\tau \cos\theta  }{z+SO\cos\tau },
$$
$$H_{dir}(\omega_{a},\omega_{b})=\frac{\tan\tau |\omega_{a}\cos\theta  +\omega_{b}\sin\theta  |}{2},$$
\begin{multline}
H_{hann}(\omega_a,\omega_b)=\nonumber\\
 \left\{ \begin{array}{ll}
\frac{\left(1 + \cos\left(\frac{\omega_a}{B_{u}}\right)\right)}{2\pi B_{u}}\left(1 + \cos\left(\frac{\omega_b}{B_{u}}\right)\right), & if  \sqrt{\omega_{a}^2+\omega_{b}^2} < \pi B_{u},\\
0,  & otherwise.
\end{array} \right.
\end{multline}
and
\begin{multline}
H_{sinc}(\omega_{a},\omega_{b})=\nonumber\\
 \left\{ \begin{array}{ll}
sinc\left(\frac{\sqrt{\omega_{a}^2+\omega_{b}^2}}{B_{v}}\right), & if \; \sqrt{\omega_{a}^2+\omega_{b}^2} < \pi B_{v}, \\
0, & otherwise.
\end{array} \right.
\end{multline}
 The formula describes the following steps:
\begin{enumerate}
\item Apply a weighting factor of $\frac{SO\cos\tau }{\sqrt{SD^2+a^2+b^2+2aSD\sin\tau \sin\theta  -2bSD\sin\tau \cos\theta  }}$ to the projection $\phi_{\theta}(a, b)$.
\item Perform the Fourier transform on the weighted projection data $\phi_{\theta}^{w}(a, b)$.
\item Conduct two-dimensional directional filtering using the 2D directional filter $H_{dir}(\omega_{a},\omega_{b})$, which incorporates a 2D von Hann window $H_{hann}$ and a 2D $sinc$ window $H_{sinc}$.
\item Apply the inverse Fourier transform to the filtered projection data.
\item Perform backprojection over all projection angles $\theta \in (0, 2\pi)$.
\end{enumerate}
\section{Real Data Experimental Results}
\begin{figure}[!htp]
	\centering
	\includegraphics[width=2.8in]{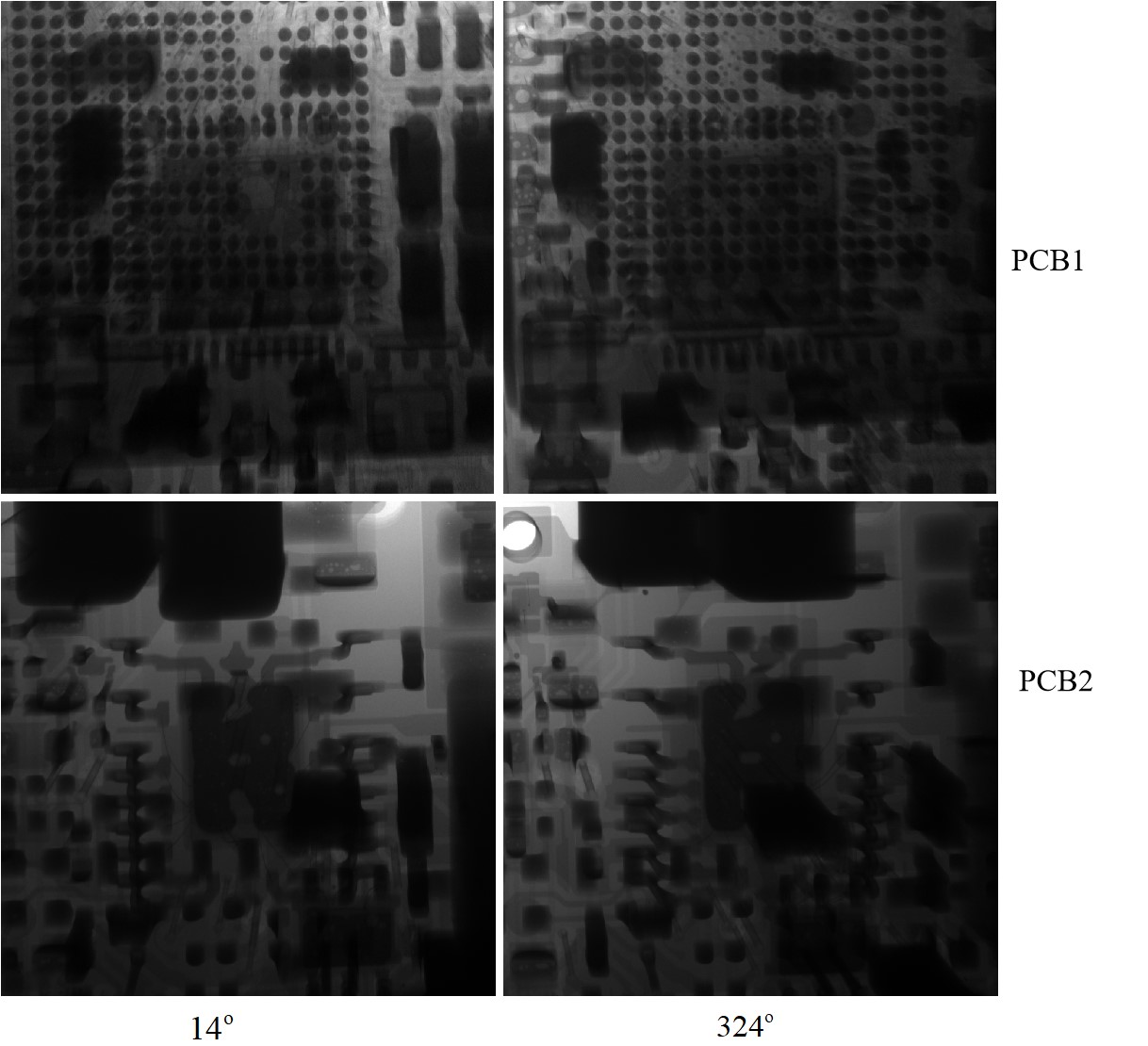}
	\caption{Cone-beam rotary CL projection data of two PCBs. The first row displays the projection data of PCB 1, while the second row corresponds to PCB 2. The first column shows the projections at an projection angle of $14^\circ$ and the second column presents the projections at  $324^\circ$.
}\label{fig2}
\end{figure}

\noindent Two actual PCBs are used to demonstrate the effectiveness of the proposed DFBP method for cone-beam Rotary Computed Laminography. The proposed DFBP method is implemented using the ASTRA-2.1 toolbox\cite{van2016fast}, and compared with the classical CL-FBP and CL-FDK methods. The projection data are acquired from a real CT scanning system, with the following geometric parameters: the source-to-rotation center is 16.12mm, the source-to-detector distance is 197.48mm, and the tilted angle is $45^\circ$. The detector array consists of $1536*1536$ elements, each with a pixel size of $0.1mm*0.1mm$. The reconstructed image volume has dimensions of 
$1280*1280*160$ voxels. In this cone-beam rotary CL scanning system, $SO\cos\tau = 16.12*\cos(45^\circ)\approx 11.3986$ and $z_{max}=0.1\times\frac{16.12}{197.48}\times\frac{160}{2}=0.6530$. Therefore, the assumption that $|z| \ll SO\cos\tau $ in \Cref{thm:5} is  satisfied.   In all experiments, the parameters are set as $B_{v} = 0.95$ and $B_{u} = 0.95$. The implementation of the three methods is carried out using MATLAB 2019b, on a computer equipped with a 3.20 GHz Intel(R) Core(TM) i7-8700 CPU and 16 GB of memory. \Cref{fig2} presents the acquired projection data from the cone-beam rotary CL system at the representative projection angles of $14^\circ$ and  $324^\circ$.
\subsection{PCB 1 Experimental Results}
\begin{figure*}[!t]
	\centering
	\includegraphics[width=0.65\textwidth]{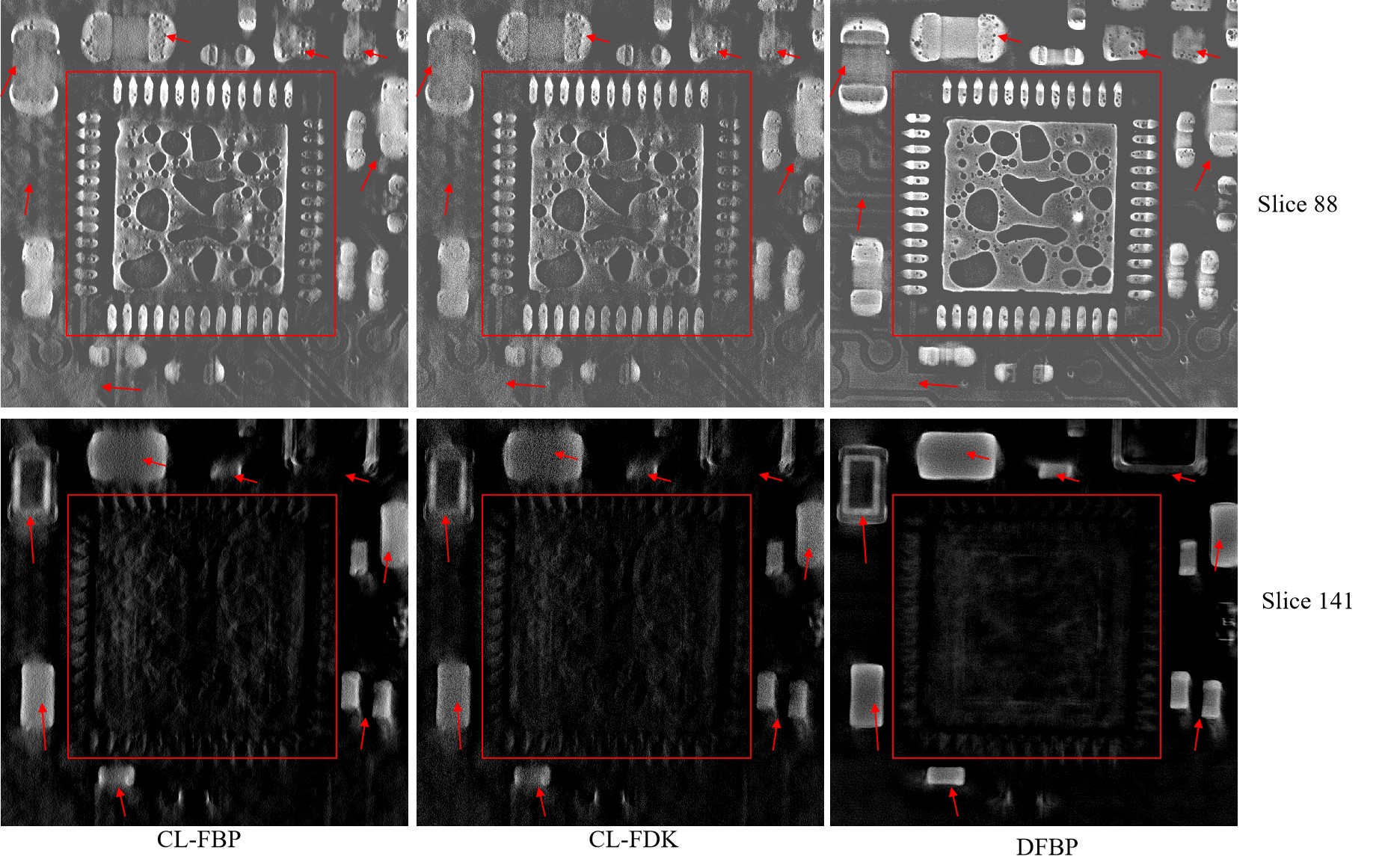}
	\caption{Comparison of image reconstruction results for PCB 1 using different algorithms. Reconstructed slices from two distinct layers (88 and 141) are displayed for the CL-FBP, CL-FDK, and the proposed DFBP. }
	\label{fig3}
\end{figure*}

\indent  \Cref{fig3} presents a qualitative comparison of the image reconstruction performance for PCB 1 using three distinct algorithms: CL-FBP, CL-FDK, and the proposed DFBP. The first and second rows display the 88th and 141st slices from the reconstructed volumes, respectively. From left to right, the columns present the results obtained using CL-FBP, CL-FDK, and the proposed DFBP. As indicated by the red arrows and  rectangles, the reconstructed results obtained using the CL-FBP and CL-FDK methods exhibit certain superposition artifacts. These artifacts, along with other observed limitations such as relatively blurred margins, reduced contrast, and noticeable smearing effects, can be primarily attributed to the fact that both methods employ conventional 1D row-wise filtering. As a result, the images suffer from a partial loss of fine structural details, particularly in the case of circuit components. Consequently, some critical defects—such as faulty solder joints and trapped air bubbles—are not fully resolved in the reconstructed images. In  contrast, the proposed DFBP algorithm mitigates these limitations through the introduction of a  2D directional filter before backprojection. As shown in the rightmost column of the \Cref{fig3}, the reconstructed results from DFBP exhibit sharper boundaries, enhanced contrast, and improved definition of fine circuit patterns. Additionally, the proposed method allows for clearer identification of subtle defects—such as micro-cracks, incomplete solder joints, and trapped air bubbles—which are relevant for PCB quality assessment. These improvements contribute to a more reliable approach for non-destructive testing.  It should be noted, however, that the proposed method introduces more  superposition artifacts in slices farther from the central layer compared to those closer to it, as indicated by the red rectangle in the lower right corner of the \Cref{fig3}. This phenomenon can be attributed to the underlying approximation condition $|z| \ll SO \cdot \cos\tau $ used in the method.
\begin{figure*}[!t]
	\centering
	\includegraphics[width=0.53\textwidth]{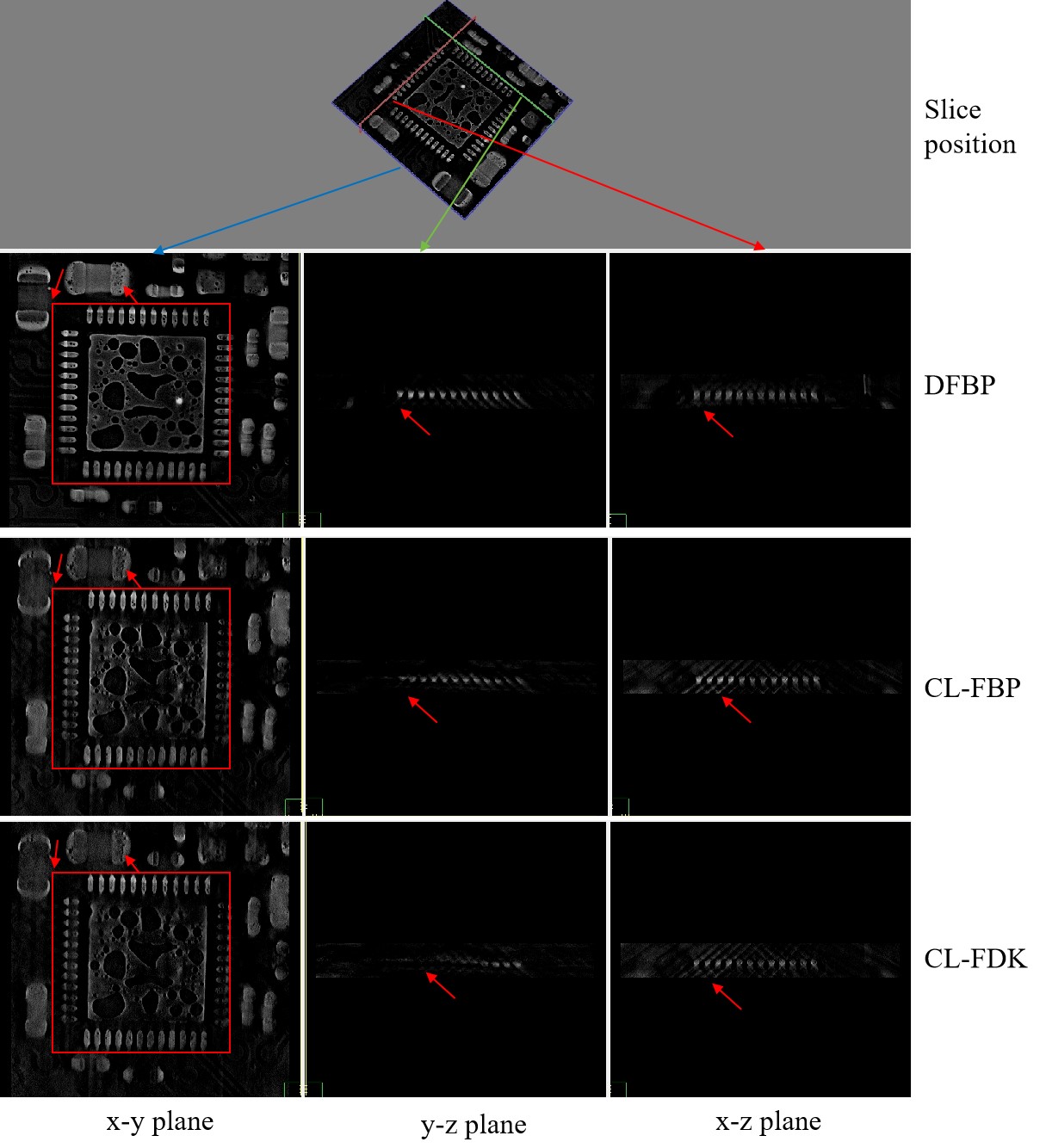}
	\caption{Cross-sectional views of different planes for PCB 1.}
	\label{fig4}
\end{figure*}
\indent \Cref{fig4}  presents cross-sectional views of various planes for PCB 1. The locations of the slices are indicated at the top, while the reconstructed results—obtained using the DFBP, CL-FBP, and CL-FDK methods—are displayed from top to bottom. From left to right, the panels correspond to the x–y, y–z, and x–z planes, respectively. \Cref{fig5} shows the volume rendering based on maximum density projection of the   results using three distinct algorithms. The above conclusion finds further validation in  \Cref{fig4} (see the red arrows and  rectangles) and \Cref{fig5}.
\begin{figure*}[!t]
	\centering
	\includegraphics[width=0.68\textwidth]{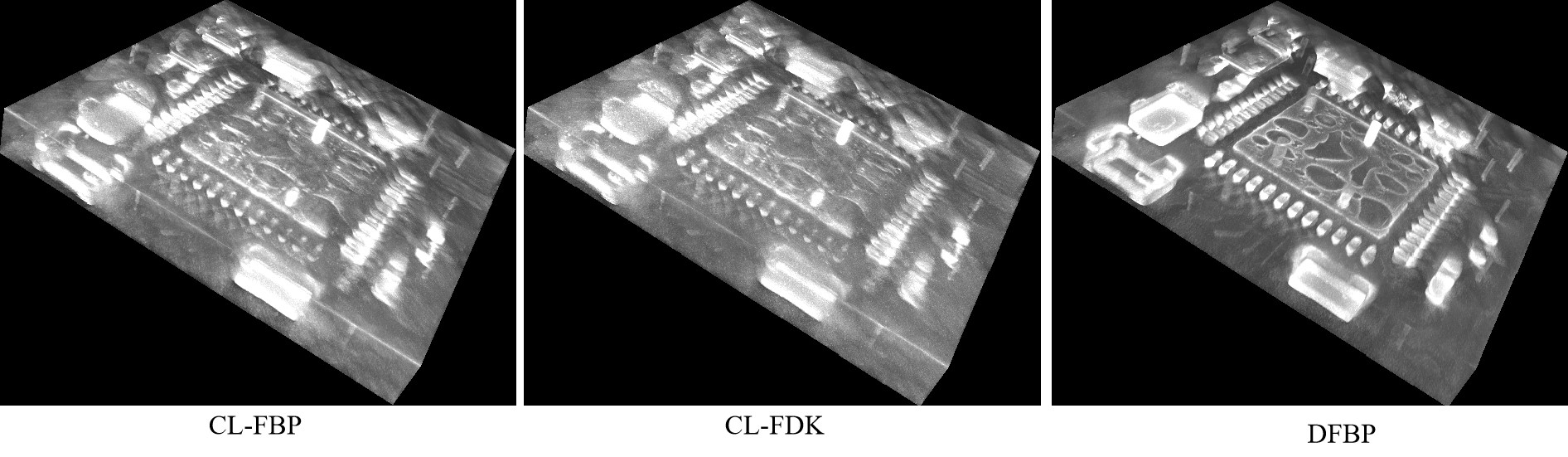}
	\caption{The volume rendering based on maximum density projection of the  reconstructed results of PCB 1 using three distinct algorithms. }
	\label{fig5}
\end{figure*}
\subsection{PCB 2 Experimental Results}
\begin{figure*}[!t]
	\centering
	\includegraphics[width=0.66\textwidth]{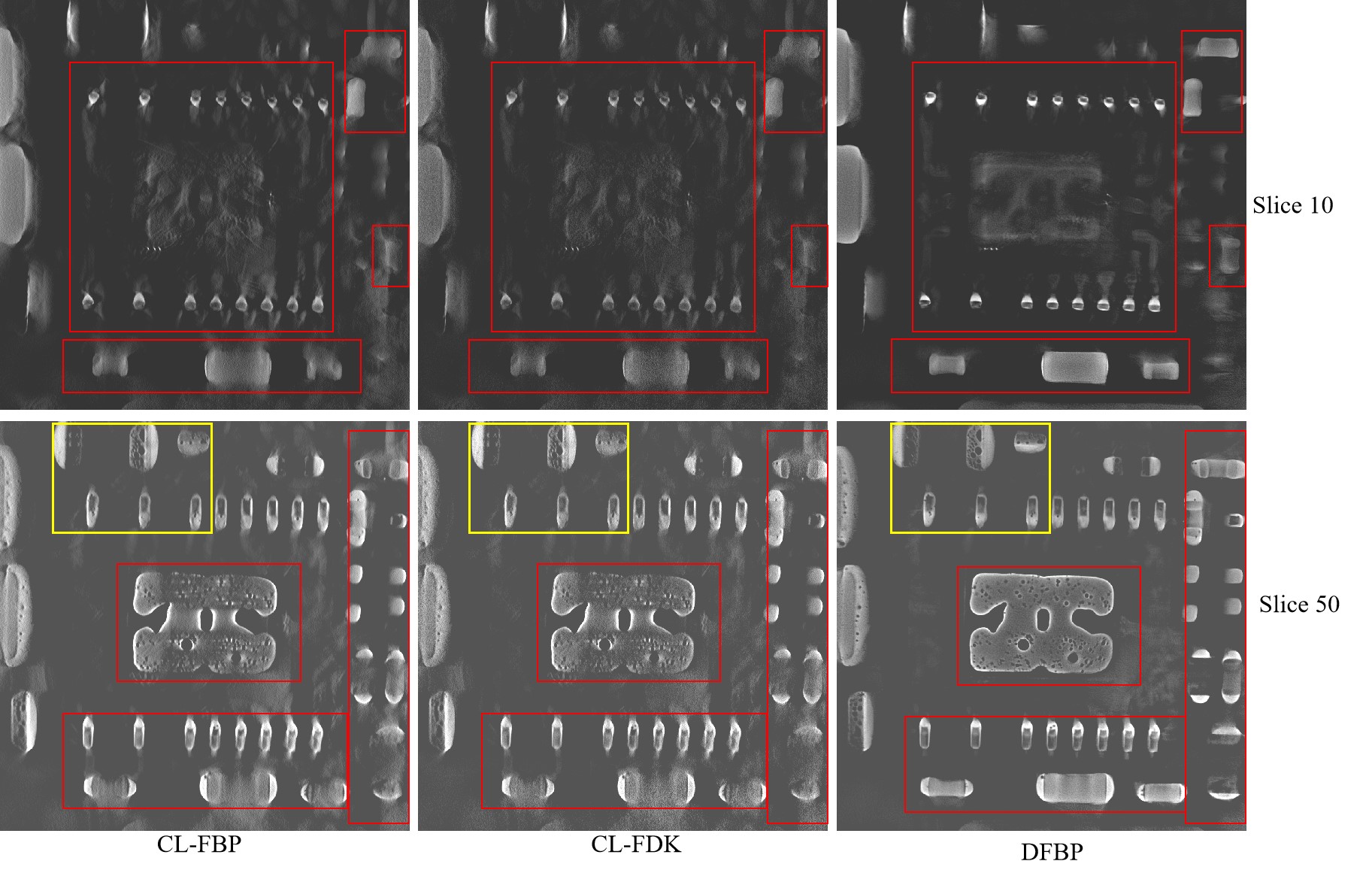}
	\caption{Comparison of image reconstruction results for PCB 2 using different algorithms. Reconstructed slices from two distinct layers (10 and 50) are displayed for the CL-FBP, CL-FDK, and the proposed DFBP.}
	\label{fig6}
\end{figure*}

\begin{figure*}[!t]
	\centering
	\includegraphics[width=0.75\textwidth]{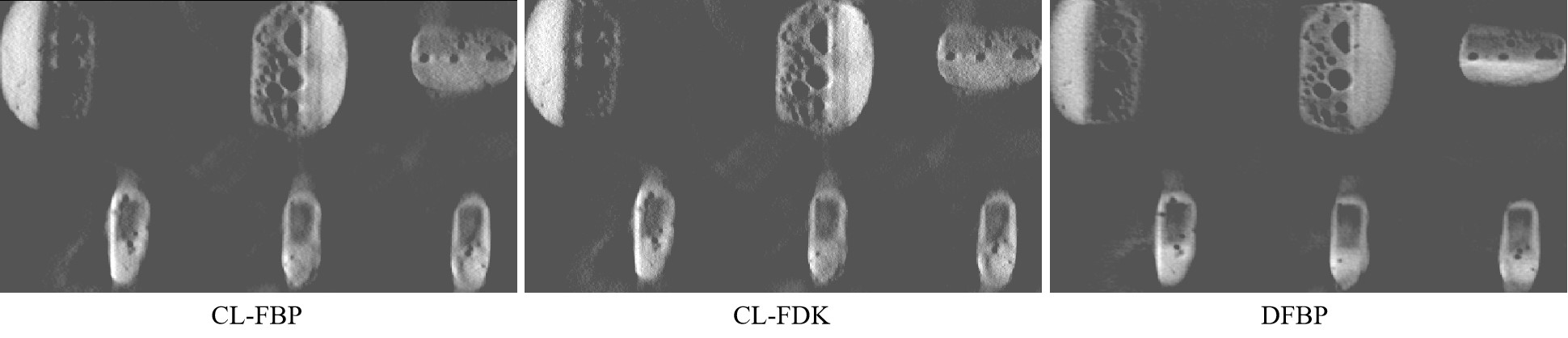}
	\caption{Magnified view of the region of interest (ROI) indicated by the yellow rectangles in \Cref{fig6}.}
	\label{fig7}
\end{figure*}

\begin{figure*}[!htp]
	\centering
	\includegraphics[width=0.58\textwidth]{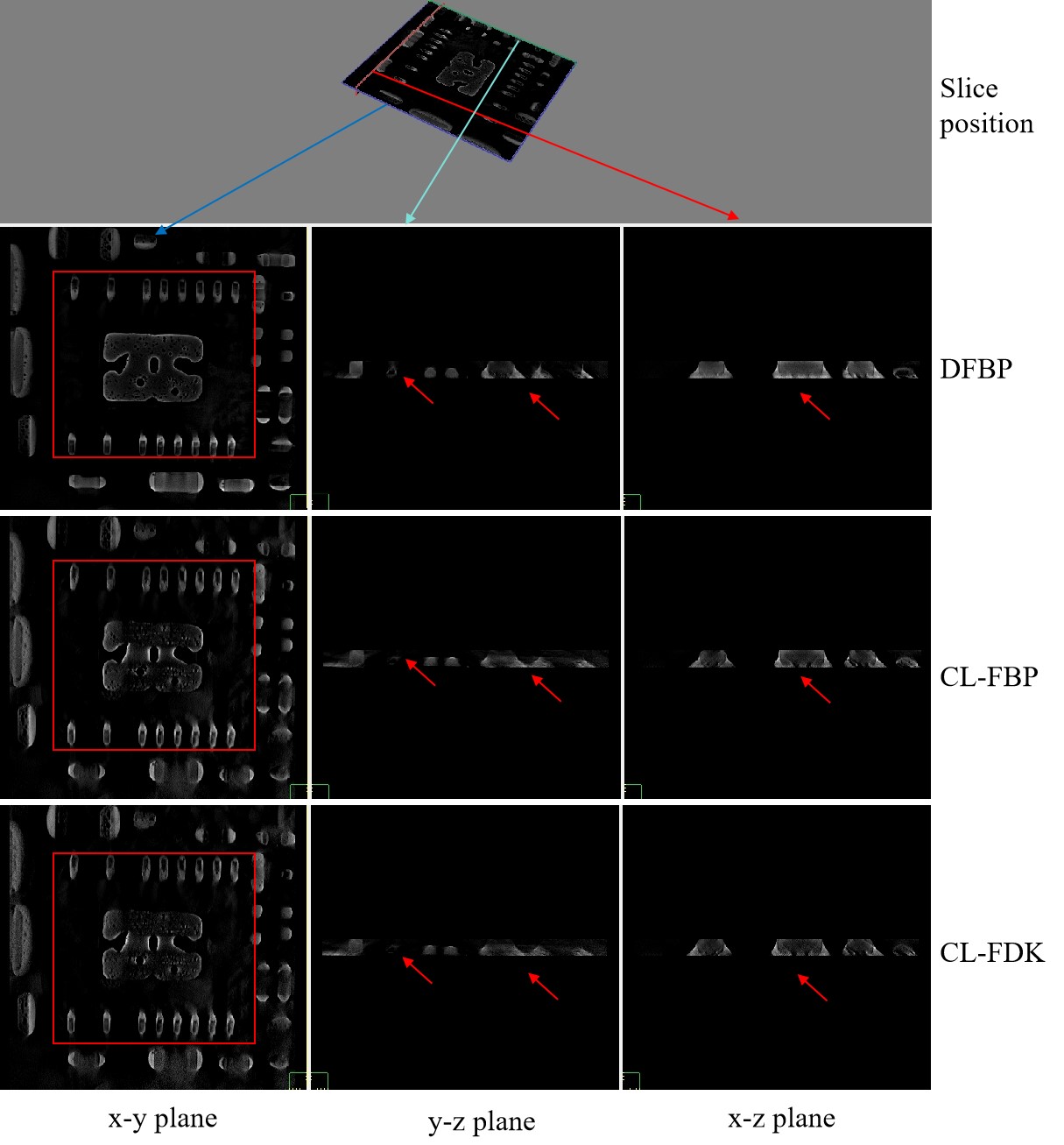}
	\caption{Cross-sectional views of different planes for PCB 2.}
	\label{fig8}
\end{figure*}

\begin{figure*}[!htp]
	\centering
	\includegraphics[width=0.75\textwidth]{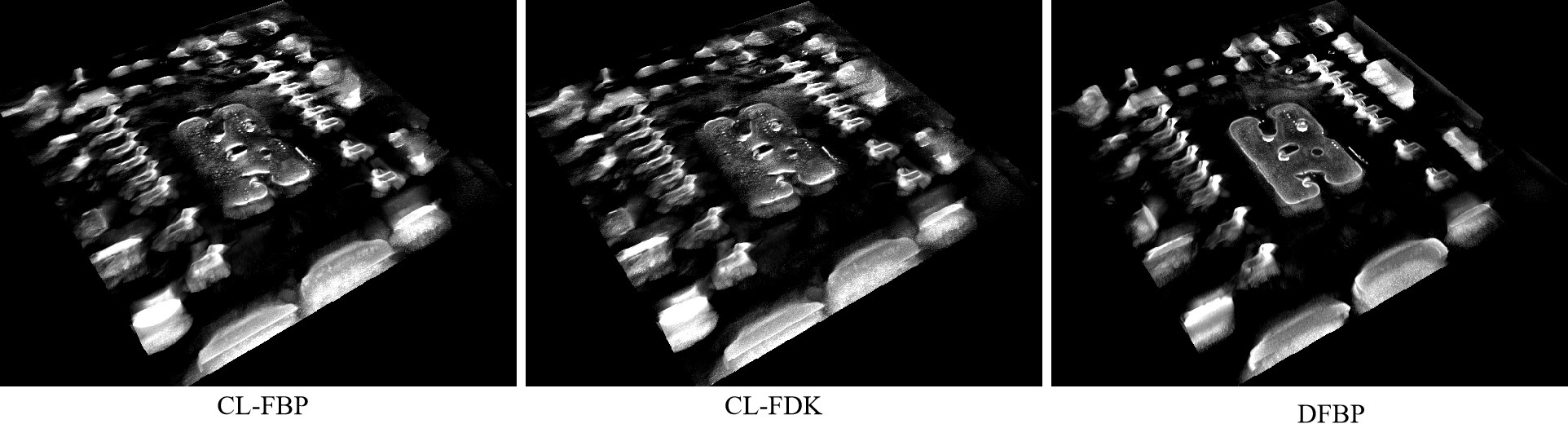}
	\caption{The volume rendering based on maximum density projection of the  reconstructed results of PCB 2 using three distinct algorithms. }
	\label{fig9}
\end{figure*}

\Cref{fig6} provides a qualitative comparison of the  reconstructed results for PCB 2, employing three different algorithms: CL-FBP, CL-FDK, and the proposed DFBP. The top and bottom rows  display  the 10th and 50th slices from the reconstructed volume, respectively. Within each column, the results from left to right correspond to CL-FBP, CL-FDK, and the proposed DFBP methods. A magnified view of the region of interest marked by the yellow rectangle in \Cref{fig6} is presented in \Cref{fig7}. As indicated by the  red rectangles in slices 10 and 50, the reconstructed results obtained using the CL-FBP and CL-FDK methods exhibit more superposition artifacts compared to those produced by the proposed DFBP method. Furthermore, the  CL-FBP and CL-FDK methods  suffer from low contrast, poorly defined structural edges accompanied by trailing artifacts (as highlighted in the red rectangles), and even fails to adequately represent bubble defects in central components, leading to blurred appearances and distorted bubble shapes (see the red rectangles in slice 50). These shortcomings result in blurred appearances and distorted bubble morphologies, significantly impairing defect recognition in non-destructive testing and subsequent quantitative analysis. In contrast, the images reconstructed by the proposed DFBP method exhibit high contrast and sharp boundaries. More notably, its advantages extend beyond these fundamental attributes to enable a more accurate and faithful reconstruction of bubble defects, as evidenced by the red rectangles in the central region of slice 50 and the  magnified view of the region of interest marked by the yellow rectangle in \Cref{fig6} is presented in \Cref{fig7}. However, it should be noted that at the 10th layer of reconstruction, our method may also introduce some superposition artifacts in the reconstructed images, which stems from our assumption  $|z| \ll SO \cdot \cos\tau $.
\\
\indent Cross-sectional views of different planes for PCB 2 are illustrated in \Cref{fig8}, where the slice positions are indicated above the \Cref{fig8}. The reconstructed results obtained via the DFBP, CL-FBP, and CL-FDK method are presented from top to bottom, with the x–y, y–z, and x–z planes displayed sequentially from left to right. In addition, \Cref{fig9} provides volume renderings based on maximum intensity projection using the three algorithms. These visual results, particularly the regions marked by red arrows and  rectangles in Figure 8, as well as the projections in Figure 9, reinforce the earlier interpretation.
\section{Conclusions and Discussions}
\noindent Cone-beam rotary CL offers a viable alternative to cone-beam CT for  objects with high aspect ratios. In such cases, conventional reconstruction methods such as CL-FBP and CL-FDK often perform poorly, as they employ 1D row-wise filtering along the detector rows. This limitation leads to reconstructed images with noticeable superimposition artifacts, reduced contrast and poorly defined structural edges. To mitigate these artifacts and enhance reconstruction quality, this paper proposes an approximate DFBP method based on the PSF for cone-beam rotary CL. This approach allows the image space filtering to be approximately transformed into operations performed on projection data. The approximate DFBP method is implemented by a three-step procedure: weighting the projection data, applying a 2D directional filter (composed of a 2D von Hann window and a 2D sinc window) to the weighted data, and finally performing the back-projection of the filtered projections.  Compared to the CL-FBP and CL-FDK methods, the proposed DFBP method yields reconstruction results with superior contrast, sharper boundaries, and fewer superposition artifacts. More notably, its advantages extend beyond these fundamental qualities to enable more accurate and faithful reconstruction of internal structures, such as bubble defects and electronic components. However, it is observed that the effectiveness of superimposition artifact suppression diminishes with increasing distance from the central layer. This can be attributed to the assumed condition $|z| \ll SO \cdot \cos\tau $ in the reconstruction model. Our future research will involve the investigation of advanced filtering techniques and the design of a tailored weighting function for cone-beam rotary CL, aimed at mitigating  artifacts.
\bibliographystyle{IEEEtran} % 使用 IEEE 官方格式
\bibliography{references} % 导入 .bib 文件（注意：无后缀名）
\begin{IEEEbiography}[{\includegraphics[width=1in,height=1.25in,clip,keepaspectratio]{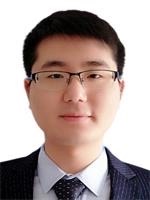}}]{Chengxiang Wang}
received the B.S. degree in Mathematics and Applied Mathematics from Sichuan University of Arts and Sciences, Dazhou, China, 
in 2011, and the Ph.D. degree in  Mathematics from Chongqing University, Chongqing, in 2016. He was a postdoctor at the School of 
Mathematical Sciences, University of Electronic Science and Technology of China, from 2017 to 2019. Now, he is an associate professor in Chongqing 
Normal University. His research interests include CT reconstruction, image processing, and optimization method.\end{IEEEbiography}
\vspace{-71pt}
\begin{IEEEbiography}[{\includegraphics[width=1in,height=1.25in,clip,keepaspectratio]{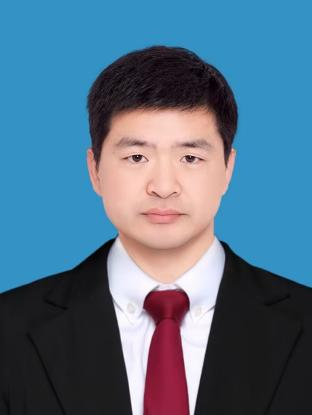}}]{Wei Yu}
received the Ph.D. degree in Instrumentation Science and Technique from Chongqing University in 2014. He did doctoral 
research at the ICT Research Center, Key Laboratory of Optoelectronic Technology of System of the Education Ministry of China 
in Chongqing University from 2011 to 2014. He is a  professor in the Hubei University of Science and Technology. His main research directions are medical imaging and  image processing.\end{IEEEbiography}
\vspace{-71pt}
\begin{IEEEbiography}[{\includegraphics[width=1in,height=1.25in,clip,keepaspectratio]{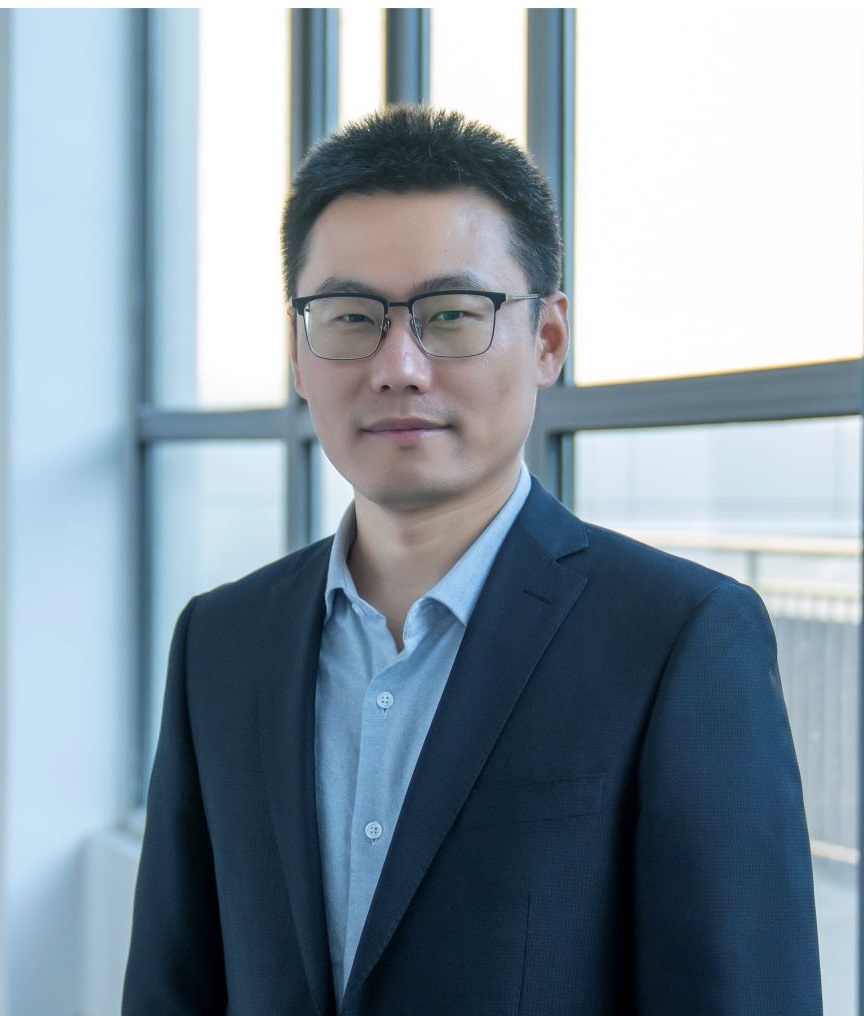}}]{Baodong Liu}
received the B.S. degree in information and computing sciences and the
Ph.D. degree in instrument science and technology from Chongqing University,
Chongqing, China,in 2005 and 2010, respectively. From 2011 to 2013, he served as a Postdoctoral Research Fellow at Wake Forest University. In 2013, he joined the Institute of High Energy Physics, Chinese Academy of Sciences, where he currently holds the position of Principal Senior Engineer. His research focuses on computed tomography and image processing.
\end{IEEEbiography}
\end{document}